\documentclass[journal]{IEEEtran}

\ifCLASSINFOpdf
\else
\fi

\usepackage[ruled,linesnumbered,vlined]{algorithm2e}
\usepackage{amsmath,amsfonts}
\usepackage[noend]{algorithmic}
\usepackage{array}
\usepackage{ragged2e}
\usepackage[caption=false,font=normalsize,
labelfont=sf,textfont=sf]{subfig}
\usepackage{textcomp}
\usepackage{stfloats}
\usepackage{url}
\usepackage{verbatim}
\usepackage{graphicx}
\usepackage{balance}
\usepackage{soul}
\usepackage{xcolor}
\sethlcolor{gray}
\usepackage{cite}
\usepackage{multirow}
\usepackage[table]{xcolor}
\usepackage{booktabs}
\usepackage{color}
\usepackage{amssymb}
\usepackage{bm}
\usepackage{amsthm}
\usepackage{siunitx}
\usepackage{adjustbox}
\usepackage{pdfpages}
\usepackage[normalem]{ulem}
\robustify\uline

\begin{document}

\title{Memetic Search for Green Vehicle Routing Problem with Private Capacitated Refueling Stations}

\author{Rui~Xu,
        Xing~Fan,
        Shengcai~Liu,~\IEEEmembership{Member,~IEEE},
        Wenjie~Chen,
        and~Ke~Tang,~\IEEEmembership{Fellow,~IEEE}
\thanks{R. Xu and X. Fan are with the School of Business, Hohai University, Nanjing 211100, China, and with the Guangdong Provincial Key Laboratory of Brain-inspired Intelligent Computation, Southern University of Science and Technology, Shenzhen 518055, China (e-mail: rxu@hhu.edu.cn; fanxingbozhou@outlook.com).}
\thanks{S. Liu and K. Tang are with the Guangdong Provincial Key Laboratory of Brain-inspired Intelligent Computation, Department of Computer Science and Engineering, Southern University of Science and Technology, Shenzhen 518055, China (e-mail: liusc3@sustech.edu.cn; tangk3@sustech.edu.cn).}
\thanks{W. Chen is with the School of Information Management, Central China Normal University, Wuhan, China.(e-mail: chenwj6@ccnu.edu.cn).}
\thanks{Corresponding author: Shengcai Liu.}
}

\markboth{IEEE TRANSACTIONS ON EVOLUTIONARY COMPUTATION,~Vol.~XX, No.~X, XX~XXXX}{Memetic Algorithm for Green Vehicle Routing Problem with Private Capacitated Stations}

\maketitle

\begin{abstract}
The green vehicle routing problem with private capacitated alternative fuel stations (GrVRP-PCAFS) extends the traditional green vehicle routing problem by considering capacitated refueling stations, where a limited number of vehicles can refuel simultaneously and additional vehicles must wait.
This feature presents new challenges for route planning, as waiting times at stations must be managed while keeping route durations within limits and reducing total travel distance.
This article presents METS, a novel memetic algorithm (MA) with separate constraint-based tour segmentation (SCTS) and a local search procedure tailored for solving GrVRP-PCAFS.
METS balances exploration and exploitation through three key components.
For exploration, the SCTS strategy splits giant tours to generate diverse solutions, and the search process is guided by a comprehensive fitness evaluation function to dynamically control feasibility and diversity to produce solutions that are both diverse and near-feasible.
For exploitation, the local search procedure incorporates tailored move operators with constant-time evaluation mechanisms, enabling efficient examination of large solution neighborhoods.
Experimental results demonstrate that METS discovers 31 new best-known solutions out of 40 instances in existing benchmark sets, achieving substantial improvements over current state-of-the-art methods.
Additionally, a new large-scale benchmark set based on real-world logistics data is introduced to facilitate future research.
\end{abstract}


\IEEEpeerreviewmaketitle

\section{Introduction}
\IEEEPARstart{A}{s} environmental concerns grow and sustainable logistics becomes increasingly critical, the green vehicle routing problem (GrVRP)~\cite{gvrp} has emerged as an essential challenge in modern transportation systems~\cite{gvrpsurvey2,CBACO,OngEV2024}.
Specifically, the GrVRP~\footnote{The acronym GrVRP is used here to avoid confusion with the classical generalized vehicle routing problem (GVRP) employed in the literature \cite{GeneVRP}.}
addresses the routing of alternative fuel vehicles (AFVs), such as electric or hydrogen-powered vehicles, by considering their limited driving range and the need for refueling while serving customers.
A good route plan should minimize travel distance while ensuring timely refueling.
In the classic GrVRP model~\cite{gvrp}, it is assumed that the alternative fuel stations (AFSs) have unlimited service spots (capacity), meaning vehicles can refuel immediately upon arrival without waiting times.

\begin{figure}[tbp]
\centering
\includegraphics[width=1\linewidth]{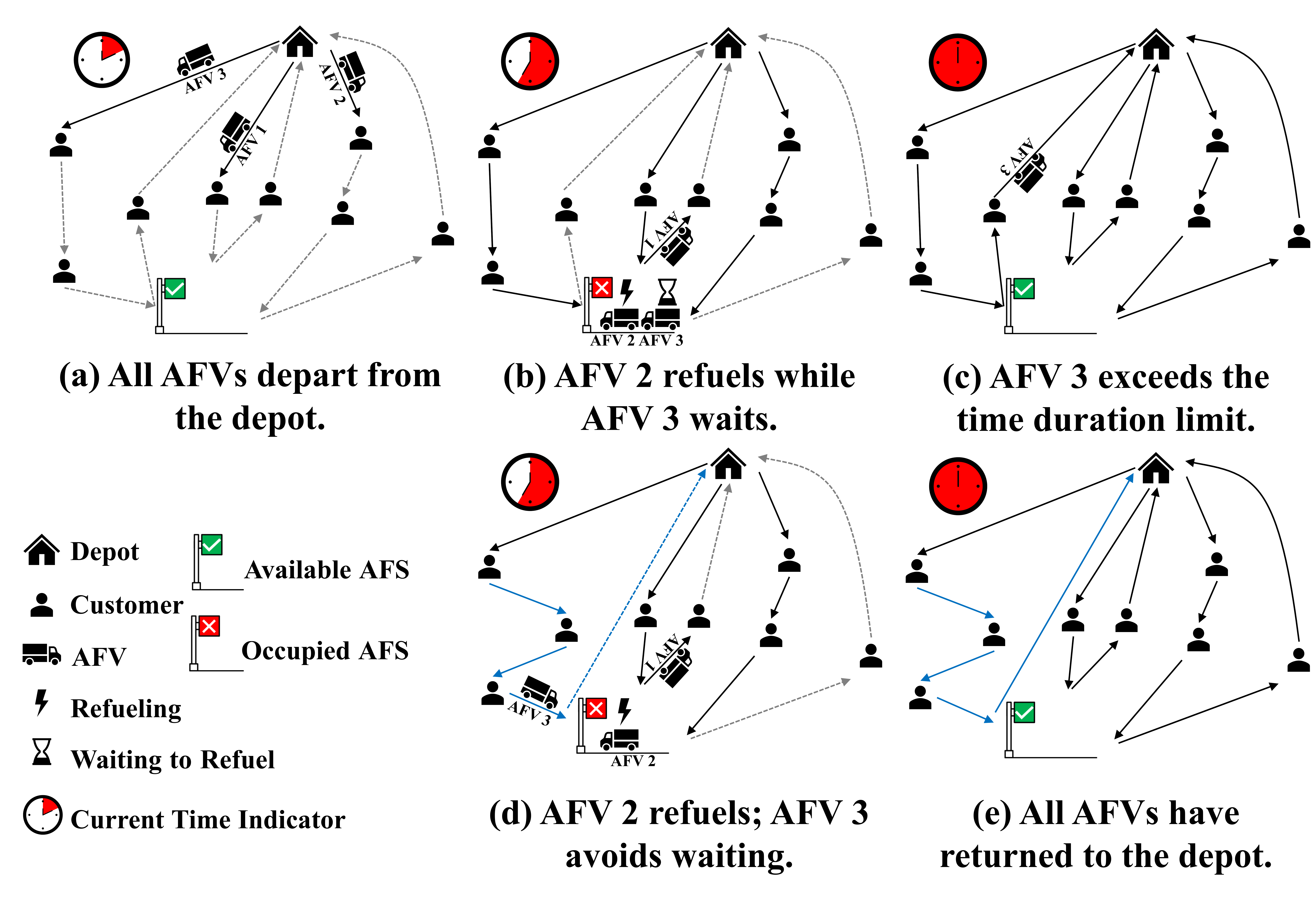}
    \caption{Illustrations of vehicle waiting caused by the limited capacity at AFS in GrVRP-PCAFS with three AFVs, eight customers, and one AFS that can serve one vehicle at the same time.
    \textbf{(a)} Three AFVs depart from the depot.
    \textbf{(b)} During service, AFV 2 is refueling while AFV 3 arrives at the same station and must wait.
    \textbf{(c)} AFS 3 exceeds the route duration limit due to waiting for refueling.
    \textbf{(d)} After adjusting the route plan (the blue line), AFS 3 serves customers first and then refuels at the AFS, thus avoiding crowding at the AFS.
    \textbf{(e)} All AFVs complete their routes within the route duration limit.}
\label{fig6}
\end{figure}

This assumption, however, does not always hold in real-world scenarios where AFSs often have a limited number of charging pumps or refueling spots~\cite{bruglieri2018,RN7153}.
Among various scenarios involving AFSs with limited capacity, Bruglieri~\textit{et~al.}~\cite{RN6218} studied the typical case where logistics companies operate their own private AFSs and introduced the GrVRP with private capacitated alternative fuel stations (GrVRP-PCAFS). 
In GrVRP-PCAFS, each AFS can only serve a limited number of vehicles simultaneously and any additional vehicle must wait until a refueling spot becomes available.
Compared to the classic GrVRP, which is already NP-hard, GrVRP-PCAFS is even more challenging to solve due to the added complexity of managing waiting times and ensuring route durations do not exceed a given time limit.
Examples of route plans for GrVRP-PCAFS are demonstrated in Fig.~\ref{fig6}, where the AFS can serve only one vehicle at the same time, and waiting times at the AFS eventually cause route duration to exceed the limit (Fig.~\ref{fig6}(b)-\ref{fig6}(c)).
Through adjusting the visiting sequence of customers and AFS in the route plan, such waiting times and route duration violations are avoided (Fig.~\ref{fig6}(d)-\ref{fig6}(e)).

Existing approaches for solving GrVRP-PCAFS can be classified into exact methods and heuristic methods.
However, neither approach has shown fully satisfactory performance.
Exact methods, such as cutting plane techniques \cite{RN6218, RN6217}, guarantee optimality but are restricted to small-scale instances, due to their exponential time complexity.
Heuristic methods, like the GRASP algorithm~\cite{RN6219}, aim to address larger instances.
However, our experiments find that for medium-scale instances with 50 and 100 customers, there is still considerable room for improvement in solution quality obtained by existing heuristic methods.
Furthermore, research on large-scale GrVRP-PCAFS instances remains limited.
The largest problem instance in the publicly available benchmark set~\cite{RN6219} contains only 100 customers.
In contrast, with the rapid growth of urban areas, a real-world GrVRP-PCAFS instance might involve many more customers.
For example, based on our collected data from a logistics company in Beijing, a major city in China, real-world applications regularly handle instances with up to 1000 customers (see Section~\ref{sec:exp} for details).
In summary, these observations indicate notable research gaps in both developing high-performing algorithms for GrVRP-PCAFS and establishing benchmark sets that can better reflect real-world problem sizes.

This work aims to address the above limitations.
Specifically, a novel \textbf{me}metic algorithm with separate constraint-based \textbf{t}our segmentation and the local \textbf{s}earch, dubbed METS, is proposed to solve GrVRP-PCAFS.
Memetic algorithms (MAs), which integrate exploration mechanisms such as crossover with exploitation-oriented local search heuristics, represent a powerful class of Evolutionary Algorithms (EAs) and have achieved state-of-the-art results on many variants of vehicle routing problems (VRPs)~\cite{haoTSP2023,VRPSPDTW3,meiyiTSP}.
However, although MA provides a generic framework, developing an effective instantiation of MA for GrVRP-PCAFS is non-trivial.
Specifically, two main challenges need to be addressed.
First, effective exploration is needed to maintain population diversity.
A common strategy is to consider both feasible and infeasible solutions during the search process.
However, it is challenging to simultaneously promote diversity and appropriately control the degree of constraint violations in GrVRP-PCAFS, since infeasible solutions with severe constraint violations, despite being diverse, cannot provide valuable solution information, e.g., patterns of customer sequences.
This is termed as the diversity-feasibility control challenge.
Second, an efficient local search procedure is essential for a high-performing MA.
It should incorporate problem-specific move operators tailored to GrVRP-PCAFS, enabling fast and effective neighborhood exploration.

METS integrates several features to comprehensively address the above two aspects.
First, for exploration, the giant-tour solution representation, which encodes a solution as a single permutation of all customers, is adopted for population initialization and crossover in METS.
To generate diverse solutions based on the giant tours, a new separate constraint-based tour split (SCTS) strategy is proposed that splits a giant tour into routes using a randomly selected single constraint, thereby producing distinct solutions.
To tackle the diversity-feasibility control challenge, a comprehensive fitness evaluation function is introduced in METS that simultaneously takes into account solution costs, constraint violation degrees, and diversity contribution.
Through adaptive adjustment of weighting parameters for these terms, METS effectively controls population diversity and feasibility during the search process.
Second, for exploitation, a local search procedure is introduced to improve solution quality. This procedure incorporates a conditional AFS Insertion (CAI) rule, which automatically determines whether to insert an AFS based on the current state of the route.
Based on this rule, four new move operators are designed to integrate AFS insertion and customer adjustment into a single operation.
Additionally, a constant-time move evaluation mechanism is proposed for these operators, significantly reducing computational overhead of the local search procedure.

The main contributions of this work are summarized below.
\begin{itemize}
 \item A novel memetic algorithm called METS is proposed for solving GrVRP-PCAFS. The algorithm has three key components: a SCTS strategy to promote solution diversity, a comprehensive fitness evaluation function for effectively controlling diversity and feasibility among population, and a local search procedure with specialized move operators tailored to GrVRP-PCAFS.
 \item Experiments show that METS discovers 31 \textbf{new} best-known solutions out of 40 instances in the existing benchmark set, demonstrating improvements by large margin over current state-of-the-art methods.
 \item A new large-scale benchmark set for GrVRP-PCAFS is established based on real-world logistics data collected from Beijing, featuring problem instances with up to 1000 customers.
 The benchmark instances and the source code of METS are open-sourced to support future research.
\end{itemize}

The rest of the article is organized as follows.
Section~\ref{sec:sec2} presents the problem description and literature review.
Section~\ref{sec:methods} first introduces the key features of METS, followed by the description of its framework.
Section~\ref{sec:exp} compares METS with existing methods on existing benchmark set and the new benchmark set introduced in this work.
Finally, conclusions and future directions are provided in  Section~\ref{sec:conclusion}.

\section{Problem Description and Related Work}
\label{sec:sec2}

\subsection{Notations and Problem Formulation}
\label{Notations and Problem Formulation}
Formally, the GrVRP-PCAFS~\cite{RN6218} is defined on a complete directed graph \( G = ( V, E) \), where \( V = \{0\} \cup V_c \cup F = \{0, 1, 2, \ldots, n, n+1, \ldots, n+s\} \) is the node set and \( 0 \) represents the depot.
\( V_c = \{1, 2, \ldots, n\} \) is a set of \( n \) customers and \( F = \{n+1, n+2, \ldots, n+s\} \) is a set of \( s \) AFSs.
The edge set \( E = \{(i, j) \mid i, j \in V, i \neq j\} \) is defined between each pair of vertices.
Each customer \( i \in V_c \) has a service time \( \tau(i) \) and each edge \( (i, j) \in E \) is associated with a travel time \( t_{i, j} \) and a travel distance \( d_{i, j} \).

The problem involves a homogeneous fleet of \( M \) AFVs initially located at the depot.
The AFVs depart from the depot with full energy level \( E_f \), serve each customer exactly once, and return to the depot.
Each AFV must adhere to a route duration limit \(T_{max}\) that includes travel time, waiting time, and service time.
This duration constraint ensures that drivers are not overworked (complying with transportation regulations) and customers are served within a reasonable period.
The AFVs consume energy at a rate \( cr \) while traveling, i.e., the maximum driving range \(D_{max}\) is determined by the full energy level \( E_f \) and the energy consumption rate, calculated as \(D_{max} = \frac{E_f}{cr}\).
An AFV must visit an AFS to refuel to the full energy level \(E_f\) before depleting its energy, with a constant refueling time \( \tau(s) \).
Following Bruglieri~\textit{et al.}~\cite{RN6218}, this work focuses on small package delivery scenarios where vehicles are assumed to have unlimited capacity.
To avoid confusion, the term ``capacity (\(\eta_s\))'' in this article refers specifically to AFSs, indicating the maximum number of vehicles that can be serviced simultaneously at each AFS.

A solution \( \varphi \) to GrVRP-PCAFS is represented by a set of vehicle routes, \( \varphi = \{r_{1}, r_{2}, . . . , r_{h}\} \), where \( h \) is the number of AFVs used.
Each route \( r_{i} \) consists of a sequence of nodes that the AFV visits, i.e., \( r_{i} = (x_{i,1}, x_{i,2}, . . . , x_{i,k_{i}}) \), where \( x_{i,j} \) is the \(j\)-th node visited in \( r_{i} \) and \( k_{i} \) is the length of \( r_{i} \).
For brevity, below we temporarily omit the subscript \( i \) in \( r_{i} \), i.e., \( r = (x_{1}, x_{2}, . . . , x_{k}) \).
The total travel distance of \( r \), \( TD(r) \), is:
\begin{align}
\label{problem eq1}
TD(r) = \sum_{j=1}^{k-1} d_{x_j,x_{j+1}}.
\end{align}

An AFV departs from the depot with full energy level \(E_f\).
The energy level of the AFV on arrival at and departure from \( x_{j} \), denoted as \( l_{x_j} \) and \( L_{x_j} \), respectively, can be computed recursively as follows:
\begin{equation}
\label{problem eq2}
\begin{aligned}
&l_{x_j} = L_{x_{j-1}} - cr \cdot d_{x_{j-1},x_j}, \quad j > 1,\\ 
&L_{x_j} =
\begin{cases} 
E_f, \quad&j = 1 \lor x_j \in F, \\
l_{x_j}, &j > 1 \land x_j \notin F. \\
\end{cases}
\end{aligned}
\end{equation}

The arrival and departure time at \( x_{j} \), denoted as \( arr_{x_j} \) and \( dep_{x_j} \), respectively, can be computed recursively as follows:
\begin{equation}
\label{problem eq3}
\begin{aligned}
&arr_{x_j} = dep_{x_{j-1}} + t_{x_{j-1},x_j}, \quad j > 1, \\
&dep_{x_j} =
\begin{cases} 
0, \quad&j = 1, \\
arr_{x_j}+\tau(x_j), &j > 1 \land x_j \notin F, \\
arr_{x_j}+\tau(s)+WT(x_j), &j > 1 \land x_j \in F. \\
\end{cases}
\end{aligned}
\end{equation}
Here, $\tau(x_j)$ and $\tau(s)$ are the service time at the customer and refueling time at the AFS, respectively.
The waiting time \(WT(x_j)\) at the AFS is determined by the $Reschedule$ procedure introduced in~\cite{RN6219}.
This procedure schedules the refueling sequence for all vehicles at the AFS while ensuring that at any time, the number of simultaneously refueling vehicles does not exceed the AFS's capacity \(\eta_s\).
If the number of vehicles requiring refueling exceeds this capacity, vehicles scheduled for later refueling slots must wait.
For details of the procedure, please refer to Appendix~A of the supplementary.
Hence, the duration of route \( r \), denoted as \( TM(r) \), is:
\begin{equation}
\label{problem eq4}
\begin{aligned}
TM(r) = arr_{x_k} - dep_{x_1} = arr_{x_k}
\end{aligned},
\end{equation}
which is the arrival time at the last visited node \(x_k\).

Finally, the objective is to find a solution  \( \varphi \) that minimizes the total travel distance \( TD(\varphi) \), as presented in Eq.~(\ref{problem eq6}):
\begin{subequations}
\label{problem eq6}
\begin{align}
&\min_{\varphi} \, TD(\varphi)  = \sum_{i=1}^h TD(r_i), \tag{5a} \\
\text{s.t.}& \;\;  x_{i,1} = x_{i,k_i} = 0, \quad 1 \leq i \leq h, \tag{5b} \\
&\; \sum_{i=1}^h \sum_{j=2}^{k_i-1} \mathbb{I}[x_{i,j} = z] = 1, \quad 1 \leq z \leq n, \tag{5c} \\
&\; 0 < h \leq M, \tag{5d}\\
&\; 0 \leq TM(r_i) \leq T_{max}, \; 1 \leq i \leq h, \tag{5e} \\
&\; 0 \leq l_{x_{i,j}} \leq L_{x_{i,j}} \leq E_f, \; 1 \leq i \leq h, \, 1 \leq j \leq k_i. \tag{5f}
\end{align}
\end{subequations}

The constraints are explained as follows:
5b) Each route must start and end at the depot;
5c) Each customer must be served exactly once (with \( \mathbb{I} \) as the indicator function);
5d) The number of used vehicles cannot exceed the available vehicles;
5e) Each route must comply with the duration limit;
5f) Each AFV must always maintain a non-negative energy level.

\subsection{Related Work}
\subsubsection{Integrated Routing and Refueling Problems}
Recent surveys on sustainable logistics and supply chain optimization have highlighted the growing need for routing strategies that jointly consider energy, environmental, and infrastructure constraints~\cite{response1,response2,response3,response4}.
Given that AFSs often have a limited number of refueling spots in practice, recent studies have explored various integrated routing and refueling models to better capture such operational constraints~\cite{response6,response7,response9}.
Bruglieri~\textit{et al.}~\cite{RN6218} introduced GrVRP-CAFS as the first  GrVRP model to consider the capacity of AFSs.
They developed both arc-based and path-based mixed integer linear programming (MILP) models, with corresponding cutting plane methods to solve problem instances with up to 15 customers optimally within acceptable computational time.
Their later work~\cite{RN6217} further improved these models to address the path cloning issues caused by multiple pumps at AFSs.
To handle larger problem instances, the same authors proposed a greedy randomized adaptive search algorithm (GRASP)~\cite{RN6219} that combines biased random construction and local search.
They also introduced the $Reschedule$ procedure with theoretical guarantees for coordinating multiple vehicles requiring refueling at the same AFS, while considering the limited capacity of the AFSs.
Their experimental results showed that GRASP could find high-quality solutions on benchmark instances with up to 100 customers, compared to previous exact methods~\cite{RN6217}.
This effective heuristic approach serves as an important reference algorithm for our study.

Besides the above studies, other researchers~\cite{RN6301,RN6228,queue1} have also investigated GrVRPs with limited AFS capacity.
In these studies, vehicle waiting times at AFSs are modeled by queuing processes that operate independently of vehicle routes.
This means route planning cannot influence the queuing dynamics at AFSs.
In contrast, GrVRP-PCAFS requires route plans to be made explicitly to reduce congestion at AFSs, thereby avoiding long waiting times.
From a practical point of view, these two types of approaches are suitable for different scenarios.
Specifically, the queuing-based approaches~\cite{RN6301,RN6228,queue1} are well-suited for modeling public refueling stations, where individual companies have limited control over the overall queuing process.
 GrVRP-PCAFS~\cite{RN6218,RN6217,RN6219} are well-suited to model scenarios involving private refueling stations, where the logistics company can coordinate all vehicle routes to manage congestion at the stations.

\subsubsection{Memetic Algorithms for Solving Vehicle Routing Problems}
As an important class of EAs, MAs have achieved success on a wide range of complex optimization problems~\cite{ong1,ong2,coello2010,coello2020,coello2023}.
Specifically, MAs have been the state-of-the-art methods for various VRPs, including the traveling salesman problem (TSP)~\cite{haoTSP2024,meiyiTSP},
capacitated VRP (CVRP)~\cite{vidalOR, RN7143},
VRP with time window (VRPTW)~\cite{huanghanVRPTW,VRPSPDTW3},
VRP with simultaneous pickup-delivery and time windows (VRPSPDTW)~\cite{VRPSPDTW1,VRPSPDTW2},
electric VRP (EVRP)~\cite{EV1,OngEV2024},
and routing problems with intermediate facilities~\cite{RN6205}.
Compared to these problems, GrVRP-PCAFS is primarily distinguished by its requirement to explicitly consider potential waiting times at AFSs to avoid route duration violations.
These early contributions provide foundational insights into the design of effective exploration–exploitation strategies in MA-based methods, such as population management~\cite{review07MAPM}, diversity control~\cite{review07MAadaptivepenalties}, and adaptive penalty mechanisms~\cite{review07MAdiversity}.
They have informed the development of our proposed METS algorithm. 
In the next section, the proposed METS algorithm is presented.

\section{Memetic Search for GrVRP-PCAFS}
\label{sec:methods}
This section first presents the three key components of METS:
the separate constraint-based tour split (SCTS) strategy,
the comprehensive fitness evaluation function,
and the local search procedure.
Then, the overall framework of METS is presented. 
The flowchart of METS is provided in Appendix~E.


\subsection{Splitting Giant Tours by the SCTS Strategy}
\label{subsec:scts}
Like many VRP studies~\cite{RN7155,Giant2,GiantReview,split3,split2018,split2019}, this study applies the giant-tour representation due to its natural compatibility with sequence-based crossover operators such as the order crossover (OX)~\cite{RN7129}.
For the giant-tour representation, a giant tour first encodes a solution as a permutation of all customers and then is split into routes to form a complete solution.
Existing splitting methods for giant tours in VRPs always consider multiple constraints simultaneously. 
While this helps generate feasible solutions, it often limits diversity and requires managing constraint interactions.

A separate constraint-based tour segmentation (SCTS) strategy is proposed, which uses a single constraint to split the giant tour in each operation. 
By decoupling multiple constraints, SCTS generates structurally diverse solutions and promotes population diversity during exploration.
In SCTS, two splitting procedures, $splitT_{max}$ and $splitD_{max}$, are developed.
Specifically, the $splitT_{max}$ procedure splits the giant tour based on the route duration limit constraint, and $splitD_{max}$ divides the giant tour considering the maximum driving range constraint. Algorithm~\ref{alg:scts} presents the details of the SCTS strategy.

\begin{algorithm}[tbp]
    \SetAlCapFnt{\footnotesize}        
    \SetAlCapNameFnt{\footnotesize}  
    \footnotesize
    \caption{ Constraint-based Tour Split (SCTS)}
    \label{alg:scts}
    \SetKwInput{KwData}{Input}
    \SetKwInput{KwResult}{Output}
    \KwData{The giant tour $C$ containing all customers, route duration limit $T_{max}$, maximum driving range $D_{max}$}
    \KwResult{A solution $\varphi_{c}$ containing a set of routes}
    \SetKw{To}{to}
    \SetKw{Append}{append}
    \SetKw{Into}{into}
    \SetKw{Break}{break}
    \SetKwProg{Fn}{Function}{:}{end}
    \SetKwFunction{Sort}{sort}
    \SetKwFunction{Normalize}{normalize}
    \SetKwFunction{Distance}{distance}
    \SetKwFunction{MSE}{MSE}
    \SetKwFunction{MLP}{MLP}
    \SetKwFunction{Len}{len}
    \SetKwFunction{InsFeature}{ins\_feature}
    \SetNoFillComment
    Generate a random number $\rho \in (0,1)$\;
    \uIf{$\rho < 0.5$} {$\varphi_c \gets$ $splitT_{max}$($C$, $T_{max}$)\;}
    \uElse{$\varphi_c \gets$ $splitD_{max}$($C$, $D_{max}$)\;}
    \Return{\textnormal{Solution} \( \varphi_c \)}
\SetKwProg{Fn}{Procedure}{:}{}
\SetKwFunction{FMain}{$splitT_{max}$}
\Fn{\FMain{$C$, $T_{max}$}}{
    $\varphi_c \gets \emptyset$, $i \gets 0$\;
    \While{$C \neq \emptyset$}{
        $i \gets i + 1$, $r_i \gets (\text{depot}, \text{depot})$\;
        \While{$C \neq \emptyset$}{
            Insert first customer $\gamma$ from $C$ before the last depot in $r_i$\;
            \uIf{$TM(r_i) < T_{max}$}{Remove $\gamma$ from $C$\;
            }
            \Else{
                Remove $\gamma$ from $r_i$, \textbf{break};
            }
            
        }
        $\varphi_c \gets \varphi_c \cup r_i$\;
    }
    \Return{\textnormal{Solution} \( \varphi_c \)}
}
\SetKwFunction{FSecondary}{$splitD_{max}$}
\Fn{\FSecondary{$C$, $D_{max}$}}{
    $\varphi_c \gets \emptyset$, $i \gets 0$\;
    \While{$C \neq \emptyset$}{
        $minAFS \gets$ the AFS nearest to the depot \; 
        $i \gets i + 1$, $r_i \gets (\text{depot}, minAFS, \text{depot})$\;
        
        \While{$C \neq \emptyset$}{
            Insert first customer $\gamma$ from $C$ before $minAFS$ in $r_i$\;
            \uIf{$TD((\textnormal{depot}, ..., minAFS)) < D_{max}$}
            {
            Remove $\gamma$ from $C$\;   
            }
            \Else
            {
                Move $\gamma$ after $minAFS$ in $r_i$\;
                \uIf{$TD((minAFS, ..., \textnormal{depot})) < D_{max}$}
                {
                Remove $\gamma$ from $C$\;
                }
                \Else
                {
                    Remove $\gamma$ from $r_i$, \textbf{break};
                }
            }
        }
        $\varphi_c \gets \varphi_c \cup r_i$\;
    }
    \Return{\textnormal{Solution} \( \varphi_c \)}
}
\end{algorithm}

In Algorithm~\ref{alg:scts},  SCTS takes the giant tour \( C \) containing all customers, the route duration limit \(T_{max}\), and the maximum driving range \(D_{max}\) as input, and returns a set of routes as the output \(\varphi_{c}\).   
A random number \( \rho \) is first generated (line 1). Based on the value of \( \rho \), one of the splitting procedures is selected to obtain the solution \( \varphi_c \), and each of them is chosen with probability 50\% (lines 2-6). 
Both of the splitting procedures return a solution \(\varphi_{c}\) (line 6).

For the procedure \( splitT_{max}\), it starts by initializing an empty set of routes \(\varphi_{c}\) and a counter \( i \) (line 8).
Then, as long as the giant tour \( C \) is not empty (line 9), a new route \( r_i \) is created as \((\text{depot}, \text{depot})\), indicating that the route starts and ends at the depot (line 10).
Next, the loop iterates to insert customers from the giant tour on condition that the giant tour \( C \) is not empty. Specifically, in each iteration, the first customer $\gamma$ from \( C \) is selected and inserted into the route \( r_i \) before the last \(\text{depot}\) (line 12).
If \(TM(r_i)\), the duration of route \( r_i \), is less than the route duration limit \(T_{max}\) (line 13), $\gamma$ is removed from \( C \) (line 14).
Otherwise, $\gamma$ is deleted from \( r_i \), and the inner loop is broken (line 15).
After that, the current route \( r_i \) is added to \( \varphi_c \) (line 17).
Once all customers from \( C \) have been inserted, the procedure returns the solution \( \varphi_c \) (line 18).

For the procedure \( splitD_{max}\), it also starts by initializing an empty set of routes \(\varphi_{c}\) and a counter \( i \) (line 20).
While the giant tour \( C \) is not empty (line 21), a new route \( r_i \) is created.
The $minAFS$ is determined as the AFS nearest to the depot (line 22).
Then, \( r_i \) is initialized as \((\text{depot}, minAFS, \text{depot})\), indicating that the route starts at the depot, visits the $minAFS$, and ends at the depot (line 23).
Then, the loop iterates to select the customers in \( C \). Each iteration selects the first customer $\gamma$ from \( C \) and adds it to the route \( r_i \) before the $minAFS$ (line 25).
For clarity, \((\textnormal{depot}, ..., minAFS)\) and \((minAFS, ..., \textnormal{depot})\) represent two segments of \( r_i \) from the depot to the $minAFS$ and from the $minAFS$ back to the depot, respectively. 
The ellipses indicate the omitted customers.
If \(TD((\textnormal{depot}, ..., minAFS))\), the travel distance of the path \((\textnormal{depot}, ..., minAFS)\), is less than the maximum driving range \(D_{max}\) (line 26), $\gamma$ is removed from \( C \) (line 27).
If \(D_{max}\) is exceeded, $\gamma$ is moved after the AFS in $r_i$, that is, in the second route path (line 29).
Next, if \(TD((minAFS, ..., \textnormal{depot}))\) is less than \(D_{max}\) (line 30), $\gamma$ is removed from \( C \) (line 31).
Otherwise, $\gamma$ is removed from $r_i$, and the inner loop stops (line 33).
The current route \( r_i \) is added to \( \varphi_c \) (line 34).
Once all customers from \( C \) have been inserted, the procedure returns the solution offspring \( \varphi_c \) (line 35). 
An example illustrating the segmentation process is included in Appendix~B of the supplementary.

\subsection{Evaluating Diversity-Feasibility by the Fitness Function}
\label{subsec:dfc}
A comprehensive fitness evaluation function is designed to jointly consider solution cost, constraint violation, and diversity contribution. 
It enables METS to adaptively control population feasibility and diversity, effectively addressing the diversity–feasibility balancing challenge during exploration.
The solution cost is measured by the total TD as defined in Eq.~(5a).
The constraint violation penalties and diversity contribution are introduced as follows.

\subsubsection{Evaluating Feasibility}
\label{subSubsec:valueFeas}
The feasibility evaluation considers three types of penalties: overtime, over-mileage, and over-capacity. 
Specifically, the overtime penalty applies to routes that exceed the maximum allowed duration \(T_{max}\). 
The over-mileage penalty penalizes any segment of a route, either before or after refueling, that exceeds the maximum driving range \(D_{max}\), ensuring that each AFV always maintains a non-negative energy level. 
The over-capacity penalty applies to stations where the number of refueling vehicles exceeds the capacity limit \(\eta_s\).

The penalties for overtime and over-mileage are computed using Eq.~(\ref{eq:Ptd}). Let \(P(r)\) denote the total penalty of overtime and over-mileage given a route \(r\). 
The parameters \( \omega^T \) and \( \omega^D \) are the penalty weights for overtime and over-mileage, respectively. \(path\) represents a segment of a route, i.e., the portion of the route that starts from a fully refueled state (either at the depot or an AFS) and ends at the next refueling stop (an AFS) or the depot.
\(TM(r)\) denotes the total duration of route \(r\) as defined in Eq.~(\ref{problem eq4}), and \(TD(path)\) denotes the travel distance of \(path\) as defined in Eq.~(\ref{problem eq1}).

\begin{align}
\label{eq:Ptd}
P(r) &= \omega^T \cdot \max\{0, TM(r) - T_{max}\} \nonumber \\
&\quad + \omega^D \cdot \sum_{path \in r} \max\{0, TD(path) - D_{max}\}
\end{align}

The first term, \(\omega^T \cdot \max\{0, TM(r) - T_{max}\}\), represents the overtime penalty, which is incurred when \(TM(r)\) exceeds \(T_{max}\).
The second term, \(\omega^D \cdot \sum_{path \in r} \max\{0, TD(path) - D_{max}\}\), represents the over-mileage penalty, which applies when the travel distance of any \(path\) exceeds \(D_{max}\). 
The summation ensures that all violating segments contribute to the total penalty.
By adjusting the penalty weights \( \omega^T \) and \( \omega^D \), the influence of these violations on the search process can be controlled. 

The penalty for over-capacity penalty is computed using Eq.~(\ref{eq:Pc}). 
Denote \(cs(s)\) as the over-capacity penalty for AFS \(s\). 
Let \(\mathcal{T}=\{q_1,q_2,...,q_{2h}\}\) record the arrival and departure times of all $h$ vehicles at the AFS, sorted in chronological order. The arrival and departure time can be calculated using Eq.~(\ref{problem eq3}). The initial waiting time \(WT\) for all vehicles is assumed to be 0. For a moment $q\in \mathcal{T}$, \(N(q)\) represents the number of vehicles refueling at AFS \(s\), and \(\Delta(q)\) represents the time interval between moment \(q\) and the next moment \(q+1\).

\begin{align}
\label{eq:Pc}
cs(s) = \sum_{q \in \mathcal{T}} \left( \max\{0, N(q) - \eta_s\} \cdot \Delta(q) \right).
\end{align}

The over-capacity penalty \( cs(s) \) for AFS \( s \) is calculated by summing the weighted penalties for all moments \( q \in \mathcal{T} \). For each moment \( q \), if the number of AFVs, \( N(q) \), exceeds \( \eta_s \), the excess \(N(q) - \eta_s\) is weighted by the time interval \( \Delta(q) \), and the total penalty is the sum of these weighted excesses.

The total penalty \(P(\varphi) \) for a solution \( \varphi \) is computed using Eq.~(\ref{eq:Pall}). 
Specifically, \( P(\varphi) \) is obtained by summing the penalties for all routes \( r \in \varphi \) and the over-capacity penalties for all AFSs \( s \in S \) in the solution. 
The penalty for each route \( P(r) \) is computed using Eq.~(\ref{eq:Ptd}), and the over-capacity penalty for each AFS is computed using Eq.~(\ref{eq:Pc}).
The parameter \( \omega^C \) represents the penalty weight for over-capacity. 

\begin{align}
\label{eq:Pall}
P(\varphi) = \sum_{r \in \varphi} {P}(r) + \omega^C\cdot \sum_{s \in S} cs(s).
\end{align}

\subsubsection{Evaluating Diversity}
\label{subSubsec:valueDiv}
The diversity contribution of individuals is evaluated by the normalized Hamming distance \(\xi(\varphi, \varphi')\) as shown in Eq.~(\ref{eq:hamming}). 
\( preA_{\varphi}(i) \) and \( preA_{\varphi'}(i) \) denote the arcs from the previous point to the customer \(i\) in individuals \(\varphi\) and \(\varphi'\), respectively. \( suA_{\varphi}(i) \) and \( suA_{\varphi'}(i) \) represent the arcs from the customer \(i\) to the next point in individuals \(\varphi\) and \(\varphi'\), respectively.
\begin{align}
\label{eq:hamming}
&\xi(\varphi, \varphi') = \frac{1}{2n} \sum_{i=1}^{n} \left[ \right. \nonumber \\
&\quad 1 \left( preA_{\varphi}({i}) \neq preA_{\varphi'}({i}) \cap preA_{\varphi}({i}) \neq suA_{\varphi'}({i}) \right) \nonumber \\
&\quad + 1 \left( suA_{\varphi}({i}) \neq suA_{\varphi'}({i}) \cap suA_{\varphi}({i}) \neq preA_{\varphi'}({i}) \right) \left. \right].
\end{align}

For each customer, the method checks whether the preceding and succeeding arcs differ between two individuals.
If they do, the difference is marked as 1.
Then, the Hamming distance $\xi(\varphi, \varphi')$ is averaged over all customers.

The diversity contribution \(\Phi(\varphi)\) of an individual \(\varphi\) is determined by calculating the average Hamming distance of \(\varphi\) to its nearest individuals in the population, as detailed in Eq.~(\ref{eq:div}).
The number of nearest individuals \(n_{close}\) is calculated as $n_{close} = nc \times n$, where $nc$ is a proportion parameter and $n$ is the number of customers.
\(\mathcal{N}_{close}\) represents the set of nearest individuals.
\begin{align}
\label{eq:div}
\Phi(\varphi) = \frac{1}{n_{close}} \sum_{\varphi' \in \mathcal{N}_{close}} \xi(\varphi, \varphi').
\end{align}
Intuitively, \(\Phi(\varphi)\) quantifies how different an individual  \(\varphi\) is from its nearest individuals in the population. 
This diversity measure helps to maintain population variety during the search process and prevents premature convergence.
\subsubsection{Fitness Evaluation Function}
\label{subSubsec:valueFitness}
To effectively assess the overall quality of a solution, we propose a comprehensive fitness evaluation function. This function integrates solution cost, constraint violation penalties, and diversity contribution.

Firstly, we explain the total quality function in Eq.~(\ref{eq:fit}). 
The total quality \(\Psi(\varphi)\) of a solution \(\varphi\) is calculated by summing the total distance \(TD(\varphi)\) and the total penalty \(P(\varphi)\).
\begin{align}
\label{eq:fit}
\Psi(\varphi) = TD(\varphi) + P(\varphi).
\end{align}
This function assesses the quality of a solution by considering both its cost (measured by total travel distance) and feasibility (penalizing constraint violations). 
A solution with a lower \(\Psi(\varphi)\) is better. 
Additionally, \(\Psi(\varphi)\) is used in local search to evaluate the quality of neighboring solutions.

Next, we introduce two ranking functions \(fit(\varphi)\) and \(dc(\varphi)\) that rank the solutions based on their total quality and diversity contribution, respectively.
\begin{align}
\label{eq:Ptdank}
fit(\varphi) = rank(\Psi(\varphi)).
\end{align}
\begin{align}
\label{eq:divrank}
dc(\varphi) = rank(\Phi(\varphi)).
\end{align}
Here, \(fit(\varphi)\) ranks the solutions in the population in ascending order of \(\Psi(\varphi)\), assigning lower values higher ranks to indicate better solutions.
For example, the solution with the smallest \(\Psi(\varphi)\) is ranked 1.
Similarly, \(dc(\varphi)\) ranks the solutions in descending order of \(\Phi(\varphi)\), giving higher ranks to solutions with greater diversity, thereby promoting population diversity.
This ranking method mitigates the effects of differing units or scales, enhancing the stability of the search process.

Finally, the comprehensive fitness function adopts the biased fitness formulation proposed by Vidal~\textit{et al.}~\cite{vidalOR}, as shown in Eq.~(\ref{eq:biasedfitness}).
It is calculated by combining two ranking functions \(fit(\varphi)\) and \(dc(\varphi)\) for solution quality and diversity contribution, respectively. 
The number of elite individuals \(nbE\) is calculated as $nbE = el \times n $, where $el$ is a proportion parameter and $n$ is the number of customers.
\(nbP\) is the total number of individuals in the population.

\begin{align}
\label{eq:biasedfitness}
BiasedFitness(\varphi) = fit(\varphi) + \left( 1 - \frac{nbE}{nbP} \right) \cdot dc(\varphi)
\end{align}

\subsection{Local Search Procedure}
\label{subsec:ls}
The local search procedure comprises two main components. First, a conditional AFS-Insert (CAI) rule is introduced, from which four new move operators are derived to effectively explore the solution neighborhoods. 
Second, a constant-time move evaluation mechanism is developed to significantly reduce the computational overhead during the search process. 
The details of these two components are provided below.

The pseudo-code of the local search procedure is presented in Algorithm~\ref{alg:ls}.
The procedure starts by setting the current solution $\varphi_{a}$ to solution $\varphi_{c}$ (line 1).
The algorithm iteratively improves the solution $\varphi_{a}$ through neighborhood exploration (lines 2–7).
It checks nine different move operators, $N_1$ to $N_9$ (see Section~\ref{Move Operators}). 
Specifically, $N_1$ to $N_4$ represent insertion and swap-based move operators following the CAI rule, and for simplicity, we denote $N_1$ to $N_4$ as CAI move operators. 
$N_5$ to $N_9$ are the classic move operators that include swap-based, 2-opt, and inter-route 2-opt* move operators.
For each $N_i$, the algorithm explores the best solution $\varphi$, starting from the current solution $\varphi_{a}$ (line 4). 
The quality $\Psi(\varphi)$ and $\Psi(\varphi_{a})$ are efficiently computed according to the constant-time move evaluation mechanism (see Section~\ref{Constant-Time Move Evaluation}). 
If $\Psi(\varphi) < \Psi(\varphi_{a})$, the algorithm updates $\varphi_{a}$ to $\varphi$ and restarts neighborhood exploration from the updated $\varphi_{a}$ (lines 5-6). 
The loop of neighborhood exploration terminates once neighborhood exploration yields no further improvement (line 7).
If $\varphi_{a}$ is infeasible and with the repair trigger rate $P_{rep}$\footnote{In this paper, we set $P_{rep} = 0.5$. The function $rand()$ returns a uniformly distributed random number in $[0,1]$.}, the Probabilistic Repair ($PRepair$) phase is triggered (line 8).
In this phase, the penalty weights are temporarily increased tenfold, and neighborhood exploration is restarted from $\varphi_{a}$ to get a new solution $\varphi_{b}$.
Otherwise, $\varphi_{b}$ is directly set to $\varphi_{a}$ (line 11).
Finally, the algorithm returns both $\varphi_{a}$ and $\varphi_{b}$ (line 12).

\begin{algorithm}[tbp]
    \SetAlCapFnt{\footnotesize}        
    \SetAlCapNameFnt{\footnotesize}  
    \footnotesize
    \caption{The Local search procedure}
    \label{alg:ls}
    \SetKwInput{KwData}{Input}
    \SetKwInput{KwResult}{Output}
    \KwData{Solution $\varphi_{c}$, total quality function $\Psi$ and repair trigger rate $P_{rep}$}
    \KwResult{Solutions $\varphi_{a}$ and $\varphi_{b}$}
    \SetKw{To}{to}
    \SetKw{Append}{append}
    \SetKw{Into}{into}
    \SetKw{Break}{break}
    \SetKwProg{Fn}{Function}{:}{end}
    \SetKwFunction{Sort}{sort}
    \SetKwFunction{Normalize}{normalize}
    \SetKwFunction{Distance}{distance}
    \SetKwFunction{MSE}{MSE}
    \SetKwFunction{MLP}{MLP}
    \SetKwFunction{Len}{len}
    \SetKwFunction{InsFeature}{ins\_feature}
    \SetNoFillComment
    $\varphi_{a}\gets \varphi_{c}$\;
    \Repeat{$\Psi(\varphi_{a}) < \Psi(\varphi)$}{
        \For{$i \gets 1$ \textnormal{\textbf{to}} $9$}
            {
            $\varphi \gets \textnormal{Best solution in } N_i \textnormal{ starting from } \varphi_a$\;
                \uIf{$\Psi(\varphi) < \Psi(\varphi_{a})$}
                    {
                    $\varphi_{a} \gets \varphi$, \textbf{break}\;
                    }
            }
    }
\uIf{$\varphi_a$ \textnormal{infeasible and} $rand() < P_{rep}$}
    {$\varphi_{b} \gets PRepair(\varphi_{a})$\;
    }
    \Else{$\varphi_{b} \gets \varphi_{a} $\;}
        
\Return{\textnormal{Solutions} $\varphi_{a}$ \textnormal{and} $\varphi_{b}$}

\end{algorithm}

\subsubsection{Move Operators}
\label{Move Operators}
Before introducing the move operators, we first introduce the conditional AFS-Insert (CAI) Rule.
The CAI Rule is a key mechanism that ensures the feasibility of move operators in the local search.
Denote $minAFS$ as the AFS closest to the customer immediately before the insertion point.
The CAI Rule inserts the $minAFS$ only when the number of customers in the route increases and the route does not  contain any AFS. This insertion mechanism is integrated with regular move operators, leading to four CAI move operators.
This conditional AFS insertion prevents situations where a solution is rejected due to excessive constraint violations. Specifically, such situations occur when the total distance $TD(\varphi)$ of a newly generated solution $\varphi$ is reduced, but the absence of an AFS results in a significant over-mileage penalty that outweighs the distance savings.

For example, consider two routes: \((x_{1}, x_{2},x_{3})\) and \((x_{4}, x_{5},x_{6}) \). 
When attempting to insert customer \( x_{1} \) after customer \( x_{4} \), a traditional move operator would produce the route \((x_{4},x_{1}, x_{5},x_{6}) \), which exceeds \( D_{max} \). 
However, with the CAI Rule, the $minAFS$ is conditionally added, producing the feasible route \((x_{4},x_{1}, minAFS, x_{5},x_{6}) \).
This rule differs from traditional AFS insertion strategies, which typically act as separate repair steps. Instead, the CAI rule is directly embedded into the local search move operators. More importantly, it evaluates whether an AFS is needed based on the current route state. As a result, CAI increases the number of feasible moves and improves the efficiency of neighborhood exploration. This effect is also confirmed by our experimental results (see Section~\ref{subsec:Ablation study}).

In local search, the move operators include CAI move operators $N_{1}$-$N_{4}$ and classic move operators $N_{5}$-$N_{9}$. 
For CAI move operators, $N_{1}$-$N_{3}$ are insertion-based, while $N_{4}$ is swap-based. 
For classic move operators, $N_{5}$-$N_{6}$ are swap-based, involving the exchange of two arcs. $N_{7}$ uses the intra-route 2-opt operator. 
$N_{8}$ and $N_{9}$ apply the inter-route 2-opt* operator, differing in the number of paths. 
After each move operator is applied, METS removes any AFS whose deletion does not result in a violation of the maximum driving range limit \( D_{max} \).

Before detailing the move operators, we first provide the following definitions. Let \(r(x)\) be the route \(r\) that includes the customer \(x\), and let \((x_i, x_j)\) denote a partial route from \( x_i \) to \( x_j \).
Denote \( y \) as the \(\alpha\)-th closest customer of \( x \) where \(\alpha = \max\{5, \lceil 5\% \cdot n \rceil\}\) and \( n \) is the number of customers \cite{RN7133}. 
Let \( x' \) and \( y' \) be the nodes following \( x \) and \( y \), respectively.

\textbf{CAI move operators $\bm{N_{1}-N_{4}}$}
\begin{itemize}
    \item {$CAI$\,-\,$Insert$ ($N_{1}$)}: Remove \(x\) and place \(x\) after \(y\). If \(r(y)\) does not include an AFS, add $minAFS$ after \(x\).
    \item {$CAI$\,-\,$Insert$\,-\,$Arc$ ($N_{2}$)}: If \(x'\) is a customer, remove \(x\) and \(x'\), then place \(x\) and \(x'\) after \(y\). When \(r(x) \neq r(y)\) and \(r(y)\) does not include an AFS, add $minAFS$ after \(x'\).
    \item {$CAI$\,-\,$Insert$\,-\,$ReversedArc$ ($N_{3}$)}: If \(x'\) is a customer, remove \(x\) and \(x'\), then place \(x'\) and \(x\) after \(y\). When \(r(x) \neq r(y)\) and \(r(y)\) does not include an AFS, add $minAFS$ after \(x\).
    \item {$CAI$\,-\,$Swap$\,-\,$Arc$ ($N_{4}$)}: If \(x'\) is a customer, swap \(x\) and \(x'\) with \(y\). When \(r(x) \neq r(y)\) and \(r(y)\) does not include an AFS, add $minAFS$ after \(x'\).
\end{itemize}
    
\textbf{Classic move operators $\bm{N_{5}-N_{9}}$}
\begin{itemize}
    \item {$Swap$ ($N_{5}$)}: Swap \(x\) and \(y\).
    \item {$Swap$\,-\,$DoubleArcs$ ($N_{6}$)}: If \(x'\) and \(y'\) are customers, swap \(x\) and \(x'\) with \(y\) and \(y\).
    \item {$2$\,-\,$opt$ ($N_{7}$)}: If \(r(x) = r(y)\), replace \((x,x')\) and \((y,y')\) with \((x,y)\) and \((x',y')\).
    \item {$2$\,-\,$opt^*$\,-\,$DoublePaths$ ($N_{8}$)}: If \(r(x) \neq r(y)\), replace \((x,x')\) and \((y,y')\) with \((x,y)\) and \((x',y')\).
    \item {$2$\,-\,$opt^*$\,-\,$TriplePaths$ ($N_{9}$)}: If \(r(x) \neq r(y)\), replace \((x,x')\) and \((y,y')\) with \((x,y')\) and \((x',y)\).
\end{itemize}

\begin{figure}[tbp]
\centering
\includegraphics[width=\linewidth]{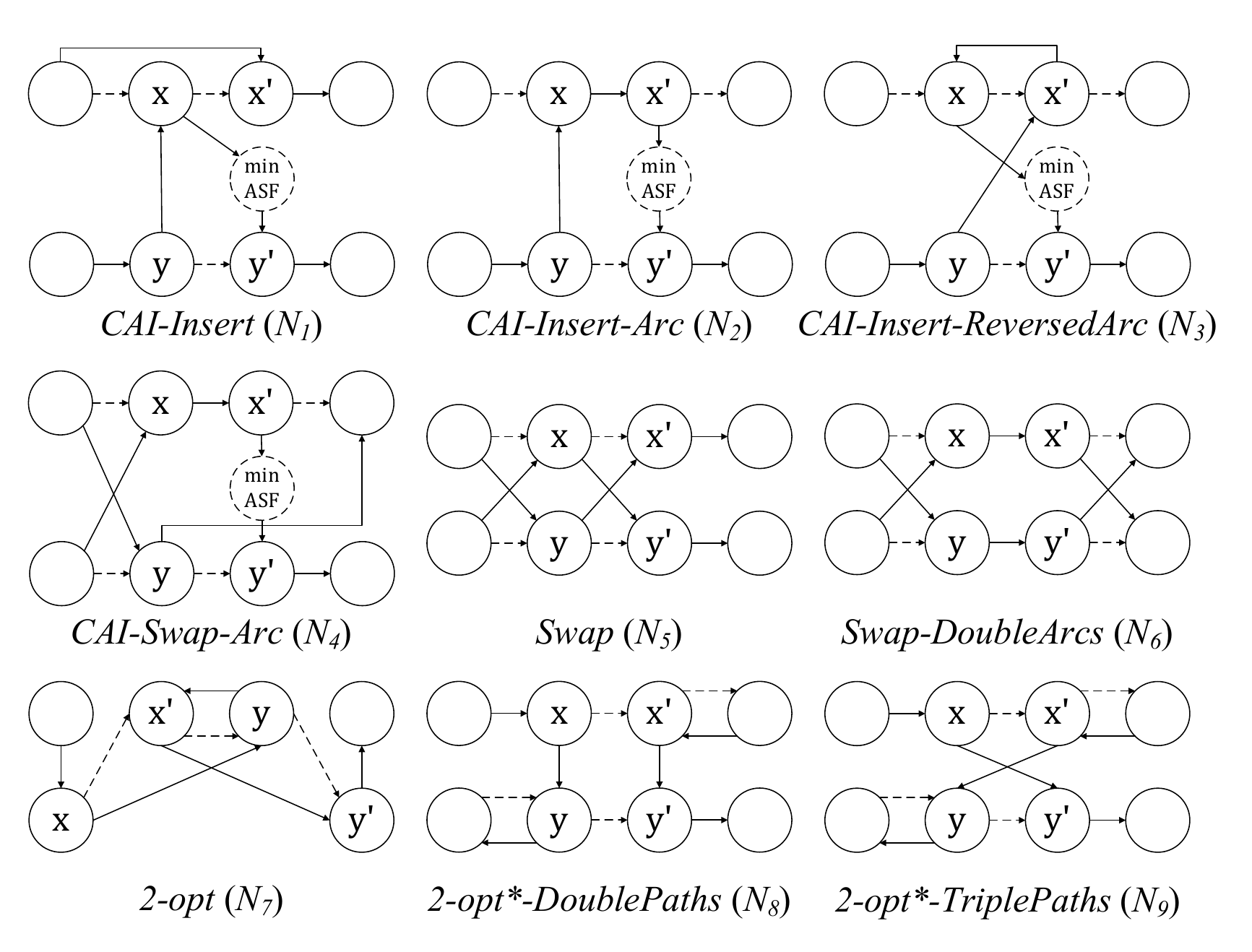}
\caption{Illustrative examples of all move operators used. Dashed lines represent the removed arcs, while solid lines indicate the new arcs added due to the move. The symbol “minAFS” represents an additional AFS inserted when the CAI rule is applied.}
\label{fig:MOVES}
\end{figure}

To provide an intuitive understanding of the employed move operators, Fig.~\ref{fig:MOVES} visually illustrates the nine move operators used in the local search procedure. 

\subsubsection{Constant-Time Move Evaluation}
\label{Constant-Time Move Evaluation}
For efficient neighborhood evaluations, we implement a fast move evaluation technique supported by a data structure. 
This technique allows neighboring solutions to be evaluated in constant time, enabling quick evaluations of routes during the local search. 
Specifically, a customer-relative-AFS (CRA) data structure is a one-dimensional array of length \(n\) to record the relative positions of customers and AFS, where \(n\) is the number of customers. 
To calculate the relative positions in a route, set \(CRA(x) = -1\) when customer \(x\) is visited before the AFS or if there is no AFS, and set \(CRA(x) = 1\) is visited after the AFS. 
\textbf{Proposition 1} presents the time complexities of evaluating neighboring solutions and exploring the complete neighborhood, with the proof provided in Appendix~F.

\textbf{Proposition 1.} In METS, for all moves operators $N_{1}$-$N_{9}$, the time complexity of evaluating neighboring solution is \(O(1)\). 
Let \(n\) be the number of customers and \(\alpha\) be the number of neighboring nodes for each customer. For each move, the time complexity of evaluating exploring the complete neighborhood is $O(\alpha n)$. 

\subsection{Overall Framework of METS}
\label{subsec:METS}

\begin{algorithm}[tbp]
\SetAlCapFnt{\footnotesize}        
\SetAlCapNameFnt{\footnotesize}    
\footnotesize                      
\caption{METS}
\label{alg:mets}
	 \SetKwInput{KwData}{Input}
    \SetKwInput{KwResult}{Output}
             \KwData{Graph $G(V,E)$, total quality function $\Psi$, lower bound of subpopulation $\mu$, upper bound of subpopulation  $\lambda$, maximum allowed number of iterations $MaxIter$, maximum allowed number of iterations without improvement $It_{N\!I}$, maximum allowed time $Maxtime$ and penalty adjust parameter $N\!S$}
            \KwResult{Best found solution $\varphi^*$}
    \SetKw{To}{to}
    \SetKw{Append}{append}
    \SetKw{Into}{into}
    \SetKw{Break}{break}
    \SetKwProg{Fn}{Function}{:}{end}
    \SetKwFunction{Sort}{sort}
    \SetKwFunction{Normalize}{normalize}
    \SetKwFunction{Distance}{distance}
    \SetKwFunction{MSE}{MSE}
    \SetKwFunction{MLP}{MLP}
    \SetKwFunction{Len}{len}
    \SetKwFunction{InsFeature}{ins\_feature}
    \SetNoFillComment
		$Initial$ $population={\{\varphi_{1},...\varphi _{p} }\} \gets SCTS(TSP) $\;
            $ \varphi^* \gets \textup{argmin} \{ \Psi(\varphi_i) , i = 1, \ldots, N_p \}$\;
            \While{$MaxIter$, $It_{N\!I}$\textnormal{, and} $Maxtime$\textnormal{ are not reached}}
            {
                Select parent solutions $\varphi_1$ and $\varphi_2$\;
                Produce a giant tour offspring $C$ from $\varphi_1$ and $\varphi_2$\;
                $\varphi_c \gets SCTS(C)$\;
                $(\varphi_{a}, \varphi_{b}) \gets Local \_Search(\varphi_c)$\;
                \uIf{$\varphi_{a} = \varphi_{b}$}{
                    Insert($\varphi_a$, \( P_{f} \), \( P_{inf} \));
                }
                \Else{
                    Insert($\varphi_a$, $\varphi_b$, \( P_{f} \), \( P_{inf} \));
                }
                Update $BiasedFitness$\;
                \uIf{\textnormal{any} \( P_i \in \{P_f, P_{inf}\} \) \textnormal{reaches} \( \lambda \)}{\( P_i \gets \textnormal{SelectSurvivors}(P_i) \);}
                \uIf{$\Psi(\varphi_{b}) < \Psi(\varphi^*)$}{$\varphi^* \gets \varphi_{b}$\;}
                \uIf{\textnormal{number of iterations} $\bmod \, NS = 0$}
                {
                    Adjust penalty parameters \( \omega^T \), \( \omega^D \) and \( \omega^C \)\;
                    Update $BiasedFitness$\;
                }
            }
            \Return{\textnormal{Best found solution} $\varphi^*$}
\end{algorithm}

The overall framework of METS is presented in Algorithm~\ref{alg:mets}.
METS begins by generating an initial population \( P \) using the $SCTS$ strategy, which splits a set of randomly generated giant tours ($TSP$).
It then records the current best solution \(\varphi^* \) (lines 1–2).
The algorithm proceeds to the main search process (lines 3–19) and continues until the termination conditions are met.
In the main search process, parent selection employs binary tournament selection based on \(BiasedFitness\) to choose \( \varphi_1 \) and \( \varphi_2 \) from both feasible and infeasible subpopulations (line 4).
Subpopulations are explicitly defined based on solution feasibility to better manage diversity and feasibility.
Specifically, \( P_f \) denotes the feasible subpopulation, while \( P_{inf} \) denotes the infeasible subpopulation.
Moreover, \(\Phi(\varphi)\), \(fit(\varphi)\) and \(dc(\varphi)\) are evaluated independently within each subpopulation (see Section~\ref{subsec:dfc}).
The Order Crossover (OX) operator \cite{RN7129} is applied to generate a giant tour \( C \) from the parent solutions (line 5). This operator randomly selects a subsequence from one parent and copies it into the offspring. The remaining positions are filled in the order they appear in the other parent, skipping any already included elements. This method preserves the relative order of nodes and tends to produce high-quality solutions.
Then, METS converts the giant tour \( C \) into a solution \(\varphi_{c}\) using the $SCTS$ strategy (line 6).

Subsequently, \(\varphi_{c}\) is refined through local search procedure, producing two improved solution \(\varphi_a\) and \(\varphi_b\) (line 7).
If \(\varphi_a = \varphi_b\), the solution is inserted into either \( P_{f} \) or \( P_{inf} \) based on its feasibility (lines 8–9).
Otherwise, both \(\varphi_a\) and \(\varphi_b\) are inserted into \( P_{f} \) or \( P_{inf} \), depending on their respective feasibility (lines 10–11).
Next, update the \( BiasedFitness \) of subpopulations (line 12).
When the size of any subpopulation \( P_i \in \{P_f, P_{inf}\} \) reaches the upper bound \(\lambda\), the \(SelectSurvivors\) procedure is triggered to maintain the subpopulation size within a reasonable range.
\(SelectSurvivors\) iteratively removes individuals, prioritizing cloned solutions and those with worse \( BiasedFitness \), until the subpopulation size reaches the lower bound \(\mu\) (lines 13–14).

If $\Psi(\varphi_{b})$ is better than $ \Psi(\varphi^*)$, \(\varphi_b\) replaces \(\varphi^* \) as the best solution (lines 15–16).
This comparison is made only between \(\varphi_b\) and the current best solution \(\varphi^* \). Since \(\varphi_b\) is either identical to \(\varphi_a\) or derived from  \(\varphi_a\) through the repair phase, it is always as good as or better than \(\varphi_a\).
Every \(NS=20\) iterations, the penalty parameters are adjusted based on the percentage of solutions that satisfy each constraint.
If any rate is \( \leq \) 15\%, indicating that the penalty is too light, the corresponding penalty is increased by 20\%. Conversely, if any rate is \( \geq \) 25\%, indicating that the penalty is too strict, it is decreased by 15\%.
The \( BiasedFitness \) is then updated according to the adjusted penalty parameters (lines 17–19).
Finally, METS returns the best feasible solution \(\varphi^* \) (line 20).

\section{Computational Studies}
\label{sec:exp}
The effectiveness of METS is evaluated through comprehensive comparisons with recent approaches for solving GrVRP-PCAFS, including the exact method CP-Proactive~\cite{RN6217}, the general-purpose solver Hexaly, and the heuristic algorithm GRASP~\cite{RN6219}, on existing public benchmark sets containing small-scale (15 customers) and medium-scale (25 to 100 customers) instances.
Additionally, a new large-scale benchmark set named Beijing is introduced, containing instances ranging from 200 to 1000 customers based on real-world logistics data collected from the city, to better reflect practical scenarios.
Both METS and GRASP are tested on the Beijing set and their results are reported.
Finally, the effectiveness of three key components in METS, i.e., the SCTS strategy, the comprehensive fitness evaluation function, and the local search procedure are also investigated.
The source code of METS and the new benchmark set are open-sourced at \url{https://github.com/FXBZ-research/METS-Algorithm}.

\subsection{Computing Environment}
All experiments were conducted on a desktop equipped with an Intel Core i5-12500H CPU at 3.1 GHz and 24 GB of RAM. METS and GRASP were implemented in Matlab (version 2022a), while the Hexaly model was implemented in Python (version 3.10.18) and solved via the Hexaly Optimizer API (version 13.5.20250610).

\subsection{Experimental Setup}
\label{sec:exp_setup}

\subsubsection{Benchmark Sets}
The experiments are conducted on three benchmark sets.
The first two sets, named CENTRAL set and Large-sized CENTRAL set, are originally proposed by Bruglieri~\textit{et al.}~\cite{RN6219}.
For clarity, these sets are renamed as S-Central and M-Central in this article.
The third set, named Beijing, is newly introduced based on real-world logistics data.
\begin{itemize}
    \item \textbf{S-Central Set.} This set contains 10 small-scale instances with 15 customers each.
    In each instance, a single AFS is located at the center of the customer area with a capacity of one AFV.
    The depot is positioned two hours away from the AFS.
    Each instance allows up to 15 available AFVs, with a route duration limit of seven hours and a maximum driving range of 160 miles.
    AFVs travel at 40 miles per hour, with both customer service time and refueling time set to 0.5 hours.
    \item \textbf{M-Central Set.} This set contains 30 medium-scale instances, divided into three groups of 10 instances each with 25, 50, and 100 customers, respectively.
    The AFS capacity increases with problem size: two AFVs for 25-customer instances, three for 50-customer instances, and eight for 100-customer instances.
    The maximum number of available AFVs is set to 7, 13, and 25 for 25, 50, and 100-customer instances, respectively.
    All instances have the same route duration limit of 7.5 hours and maintain the same vehicle speed, service time, and refueling time as the S-CENTRAL set.
    \item \textbf{Beijing Set.} This new set is created based on data from the JD Logistics company in Beijing, China.
    The original collected data contains 3000 delivery requests over several days in the city.
    From this dataset, 20 large-scale instances are generated through sampling, with customer sizes being 200, 400, 600, 800, and 1000 (4 instances per size).
    The AFS capacity increases with problem size: 20, 40, 60, 80, and 100 AFVs for 200, 400, 600, 800, and 1000-customer instances, respectively.
    The locations of customers, depot, and AFSs are specified using latitude and longitude coordinates, and the distances between them are calculated as the Euclidean distance. 
    The route duration limit is set to 8 hours, with no strict limit on the number of available AFVs.
    The vehicle speed, service time and refueling time remain the same as the previous two sets.
\end{itemize}

\begin{table}[t]
\caption{Description and Ranges of the Parameters of METS Used for Automatic Parameter Tuning with Irace~\cite{RN6323}.}
\setlength{\tabcolsep}{3pt} 

\begin{center}
\scalebox{0.88}{
\begin{tabular}{l p{3.9cm} l l l}
\toprule
\textbf{Parameter} & \textbf{Description} & \textbf{Type} & \textbf{Value Range} & \textbf{Value} \\ 
\midrule
$\omega^T$ & Overtime penalty parameter & Integer & [1, 1000] & 527 \\ 
$\omega^D$ & Over-mileage penalty parameter & Integer & [1, 1000] & 430 \\ 
$\omega^C$ & Over-capacity penalty parameter & Integer & [1, 1000] & 195 \\ 
$\mu$ & lower bound of subpopulation size & Integer & [5, 200] & 154 \\ 
$\lambda$ & upper bound of subpopulation size & Integer & [10, 400] & 222 \\ 
$el$ & Proportion of elite individuals  & Real &  (0,1)  & 0.5 \\ 
$nc$ & Proportion of close individuals  & Real & (0.1) & 0.2 \\ 
\bottomrule
\end{tabular}}
\end{center}
\label{tab1}
\end{table}

\subsubsection{Compared Methods}
METS is compared with the state-of-the-art methods for solving GrVRP-PCAFS, including the exact method CP-Proactive~\cite{RN6217}, heuristic algorithm GRASP~\cite{RN6219}, and the general-purpose global optimization solver Hexaly~\cite{hexalyref} for solving routing problems.
Compared to CP-Proactive that is restricted to small-scale instances, GRASP has better scalability as it can find high-quality solutions for medium-scale instances within reasonable computational time. Moreover, Hexaly could only find feasible solutions for small-scale instances. Hence, GRASP is the primary compared algorithm in the experiments.

As the original GRASP code is unavailable, both algorithms are implemented in Matlab for fair comparison.
The original GRASP implementation used a 2-minute time limit as the termination condition.
We find that for instances with 15 and 25 customers, GRASP typically converges well before this time limit.
However, for instances with 50 and 100 customers, GRASP is far from convergence within 2 minutes.
Therefore, to ensure sufficient convergence, both GRASP and METS are set to terminate when the maximum number of iterations reaches 2000 or when 300 consecutive iterations show no improvement.

For small and medium-scale instances, both METS and GRASP were executed 30 independent times, and their best and average solution quality as well as average time-to-best are reported. For large-scale instances in the Beijing set, due to the extensive computational requirements, both algorithms are executed 10 independent times.
A time limit of 3600 seconds was applied to both Hexaly and CP-Proactive, consistent with the original study~\cite{RN6217}.

\subsubsection{Parameter Settings}

For parameter settings in METS, the population management parameters $el$ and $nc$ are set to 0.5 and 0.2, respectively, based on recommended settings from existing EAs for solving VRPs that incorporate similar population control mechanisms~\cite{vidalOR}. 
The remaining parameters are then tuned using the automatic parameter tuning tool Irace~\cite{RN6323}. 
The training set consists of 10 randomly selected problem instances from all 60 test instances, with the tuning budget (maximum number of the algorithm runs during tuning) set to 2000.
The parameters and their final values are summarized in Table~\ref{tab1}.
The number of subpopulations ranges from $\mu$ to $\lambda$.
These values are also recommended for future research when METS is employed, since METS with such setting demonstrates good performance across all three benchmark sets in the experiments.
For fair comparison, GRASP also undergoes the same tuning procedure for its parameter $\beta$, which influences the solution construction process.
The tuned value for $\beta$ is 0.23, which is then used in the experiments.

\begin{table*}[tbp]
\caption{Comparisons between METS and reference algorithms on small-scale instances in the S-Central set.
For each instance, the best performance is indicated in bold, \textbf{new} best-known solutions (BKSs) are highlighted in gray, and results significantly better than GRASP, based on 30 independent runs, are marked with underlines according to the Wilcoxon signed rank test at a significance level of p-values \( < \) 0.05.}
\begin{center}
\scalebox{0.8}{ 
\begin{tabular}{l r r
                r r l r r
                l l r r
                r}
\toprule
 \multirow{2}{*}{Instance}   
&\multirow{2}{*}{CP-Proactive }  
&\multirow{2}{*}{Hexaly }  
& \multicolumn{5}{l}{ \multirow{2}{*}{GRASP}}                                        
& \multicolumn{4}{l}{ \multirow{2}{*}{METS}}                                  

& \multicolumn{1}{l}{Imp. A}\\ 
& \multicolumn{6}{l}{ {}}                        
& \multicolumn{5}{l}{ {}}                     
& \multicolumn{1}{c}{(\%)} \\ 
              \cmidrule(lr){2-2}       \cmidrule(lr){3-3}\cmidrule(lr){4-8}                                                             \cmidrule(lr){9-12}
              & Best~cite{RN6217}     & Best           & Best\!\cite{RN6219} & Best            & Avg$\pm$Std      & Worst   & Time (s) & Best($\Delta$BKS)                      & Avg$\pm$Std  & Worst  & Time (s) &          \\
\midrule
S-Central\_1  & \textbf{953.94}        & \textbf{953.94}& \textbf{953.94}     & \textbf{953.94} & 955.85$\pm$5.23  & 972.50  & 8.15     & \textbf{953.94}                           &\uline{953.94$\pm$0.00}  &953.94  & 17.63    & -0.20\% \\
S-Central\_2  & \textbf{948.69}        & 1062.73        & 959.88              & 959.88          & 963.95$\pm$2.39  & 968.98  & 23.05    & 959.88                                    &\uline{959.88$\pm$0.00}  &959.88  & 16.24    & -0.42\% \\
S-Central\_3  & \textbf{943.12}        & 958.94         & 958.94              & 958.94          & 967.09$\pm$6.71  & 981.85  & 28.09    & 958.94                                    &\uline{959.24$\pm$0.40}  &959.83  & 19.43    & -0.81\% \\
S-Central\_4  & 967.96                 & 1098.49        & 1099.24             & 1098.49         & 1128.93$\pm$17.06 & 1167.85 & 56.37   & \cellcolor{gray}{\textbf{947.98(-19.98)}} &\uline{1086.91$\pm$37.92}&1099.24 & 27.66    & -3.72\% \\
S-Central\_5  & \textbf{714.55}        & \textbf{714.55}& \textbf{714.55}     & \textbf{714.55} & 714.55$\pm$0.00  & 714.55  & 1.74     & \textbf{714.55}                           &714.55$\pm$0.00        &714.55  & 9.68     & 0.00\%  \\
S-Central\_6  & \textbf{844.43}        & \textbf{844.43}& \textbf{844.43}     & 845.53          & 863.22$\pm$12.32 & 886.56  & 2.29     & \textbf{844.43}                           &\uline{844.43$\pm$0.00}  &844.43  & 17.35    & -2.18\% \\
S-Central\_7  & \textbf{862.68}        & \textbf{862.68}& \textbf{862.68}     & 867.78          & 888.24$\pm$12.99 & 905.44  & 1.76     & \textbf{862.68}                           &\uline{862.68$\pm$0.00}  &862.68  & 14.59    & -2.88\% \\
S-Central\_8  & \textbf{712.83}        & \textbf{712.83}& \textbf{712.83}     & \textbf{712.83} & 723.29$\pm$9.86  & 742.61  & 1.14     & \textbf{712.83}                           &\uline{712.83$\pm$0.00}  &712.83  & 21.05    & -1.45\% \\
S-Central\_9  & \textbf{855.43}        & \textbf{855.43}& \textbf{855.43}     & \textbf{855.43} & 874.88$\pm$12.09 & 896.99  & 1.34     & \textbf{855.43}                           &\uline{855.43$\pm$0.00}  &855.43  & 1.89     & -2.22\% \\
S-Central\_10 & \textbf{901.19}        & 906.76         & 905.59              & 905.59          & 924.57$\pm$10.89 & 946.61  & 2.43     & 905.59                                    &\uline{906.23$\pm$0.56}  &906.76  & 11.35    & -1.98\% \\
\midrule                                                                                                                                                                                                       
No. Best / W-D-L & 9                    & 6              & 6                   & 4               &                  &         &          & 7                                         & 9-1-0                   &        &          &         \\
Mean           & 870.48                & 897.08         & 886.75              & 887.30          & 900.46$\pm$8.14  & 918.39  & 12.64    & 871.63                                    & 885.61$\pm$3.89         &886.96  & 15.69    & -1.59\% \\
\bottomrule
\end{tabular}
\label{tab:small_scale}
}
\end{center}
\end{table*}

\begin{table*}[h]
	\caption{Comparisons between GRASP and METS on medium-scale instances. For each instance,  the best performance is indicated in bold, new best-known solutions are highlighted in gray, and results significantly better than GRASP, based on 30 independent runs, are marked with underlines according to the Wilcoxon signed rank test at a significance level of p-values\( < \) 0.05.}
	\begin{center}
		\scalebox{0.88}{ 
			\begin{tabular}{l 
					r r l r r
					l l r r 
					r}
				\toprule
				\multirow{2}{*}{Instance}              
				& \multicolumn{5}{l}{ \multirow{2}{*}{GRASP}}                   
				& \multicolumn{4}{l}{ \multirow{2}{*}{METS}}                     
				
				& \multicolumn{1}{l}{Imp. A}   
				\\ 
				& \multicolumn{5}{l}{ {}}                        
				& \multicolumn{4}{l}{ {}}                     
				
				& \multicolumn{1}{c}{(\%)}\\ 
				\cmidrule(lr){2-6}                               \cmidrule(lr){7-10}
				&Best\!\cite{RN6219} &Best      &Avg$\pm$Std         & Worst &  Time (s)     & Best($\Delta$BKS)                          & Avg$\pm$Std               &Worst     &Time (s)&          \\ 
				\midrule
				M-Central25\_1             & 1129.85  & 1135.96  & 1152.87$\pm$28.17  &1297.96& 30.59         & \cellcolor{gray}{\textbf{1129.71(-0.14)}}  & \uline{1129.71$\pm$0.00}  &1129.71   & 34.02  & -2.01\%  \\
				M-Central25\_2             & 1113.97  & 1118.35  & 1136.12$\pm$7.39   &1147.86& 21.73         & \cellcolor{gray}{\textbf{1113.80(-0.17)}}  & \uline{1113.80$\pm$0.00}  &1113.80   & 22.64  & -1.96\%  \\
				M-Central25\_3             & 1321.82  & 1324.64  & 1344.36$\pm$14.82  &1372.72& 126.47        & \cellcolor{gray}{\textbf{1320.27(-1.55)}}  & \uline{1320.27$\pm$0.00}  &1320.27   & 4.88   & -1.79\%  \\
				M-Central25\_4             & 1122.73  & 1126.69  & 1140.37$\pm$7.46   &1155.15& 30.23         & \cellcolor{gray}{\textbf{1118.87(-3.86)}}  & \uline{1121.38$\pm$1.25}  &1122.71   & 34.51  & -1.67\%  \\
				M-Central25\_5             & 1110.36  & 1122.85  & 1140.74$\pm$7.90   &1154.31& 24.47         & \cellcolor{gray}{\textbf{1109.55(-0.81)}}  & \uline{1109.57$\pm$0.05}  &1109.73   & 20.37  & -2.73\%  \\
				M-Central25\_6             & 1090.04  & 1092.91  & 1118.17$\pm$12.65  &1143.58& 7.29          & \cellcolor{gray}{\textbf{1089.92(-0.12)}}  & \uline{1089.92$\pm$0.00}  &1089.92   & 4.10   & -2.53\%  \\
				M-Central25\_7             & 1105.04  & 1112.69  & 1133.56$\pm$10.74  &1155.24& 17.98         & \cellcolor{gray}{\textbf{1103.82(-1.22)}}  & \uline{1103.82$\pm$0.00}  &1103.82   & 15.79  & -2.62\%  \\
				M-Central25\_8             & 1141.23  & 1137.81  & 1216.42$\pm$82.64  &1335.61& 31.93         & \cellcolor{gray}{\textbf{1134.93(-6.30)}}  & \uline{1137.60$\pm$1.38}  &1139.85   & 42.67  & -6.48\%  \\
				M-Central25\_9             & 1281.57  & 1288.38  & 1303.99$\pm$12.55  &1337.88& 30.11         & \cellcolor{gray}{\textbf{1275.57(-6.00)}}  & \uline{1280.19$\pm$1.88}  &1281.43   & 20.03  & -1.83\%  \\
				M-Central25\_10            & 1311.67  & 1320.85  & 1337.36$\pm$12.10  &1375.34& 53.88         & \cellcolor{gray}{\textbf{1311.53(-0.14)}}  & \uline{1311.53$\pm$0.00}  &1311.53   & 3.48   & -1.93\%  \\
				
				M-Central50\_1             & 2520.33  & 2498.19  & 2529.29$\pm$13.79  &2556.52& 587.29        & \cellcolor{gray}{\textbf{2441.41(-78.92)}}  & \uline{2461.88$\pm$9.95}  &2487.88   & 94.88  & -2.66\%  \\ 
				M-Central50\_2             & 2435.78  & 2413.48  & 2461.84$\pm$19.42  &2492.10& 1112.44       & \cellcolor{gray}{\textbf{2241.35(-194.43)}} & \uline{2345.87$\pm$54.02} &2382.91   & 116.12 & -4.71\%  \\ 
				M-Central50\_3             & 2425.64  & 2268.80  & 2411.91$\pm$74.39  &2491.09& 620.75        & \cellcolor{gray}{\textbf{2230.13(-195.51)}} & \uline{2239.30$\pm$4.18}  &2245.35   & 153.98 & -7.16\%  \\ 
				M-Central50\_4             & 2271.14  & 2257.60  & 2332.61$\pm$78.14  &2460.95& 426.97        & \cellcolor{gray}{\textbf{2193.74(-77.40)}}  & \uline{2200.93$\pm$6.13}  &2225.82   & 116.19 & -5.65\%  \\
				M-Central50\_5             & 2467.29  & 2453.61  & 2490.11$\pm$16.55  &2521.61& 686.60        & \cellcolor{gray}{\textbf{2393.32(-73.97)}}  & \uline{2403.36$\pm$5.24}  &2412.53   & 66.45  & -3.48\%  \\ 
				M-Central50\_6             & 2422.87  & 2446.28  & 2472.57$\pm$17.07  &2512.96& 583.80        & \cellcolor{gray}{\textbf{2380.99(-41.88)}}  & \uline{2388.86$\pm$3.73}  &2395.75   & 114.55 & -3.39\%  \\
				M-Central50\_7             & 2405.19  & 2262.98  & 2414.30$\pm$73.79  &2504.46& 777.76        & \cellcolor{gray}{\textbf{2221.61(-183.58)}} & \uline{2241.80$\pm$8.54}  &2258.02   & 169.11 & -7.14\%  \\
				M-Central50\_8             & 2487.32  & 2474.34  & 2518.88$\pm$18.16  &2546.37& 779.13        & \cellcolor{gray}{\textbf{2415.43(-71.89)}}  & \uline{2421.01$\pm$4.79}  &2435.70   & 149.28 & -3.89\%  \\
				M-Central50\_9             & 2450.04  & 2409.39  & 2460.41$\pm$19.14  &2501.71& 648.02        & \cellcolor{gray}{\textbf{2241.03(-209.01)}} & \uline{2370.72$\pm$31.84} &2385.82   & 117.36 & -3.65\%  \\
				M-Central50\_10            & 2473.63  & 2436.74  & 2484.88$\pm$21.40  &2520.89& 682.37        & \cellcolor{gray}{\textbf{2402.03(-71.60)}}  & \uline{2407.71$\pm$2.60}  &2413.51   & 100.30 & -3.11\%  \\
				
				M-Central100\_1            & 4966.03  & 4766.64  & 4932.28$\pm$87.24  &5029.19& 1281.24       & \cellcolor{gray}{\textbf{4645.27(-320.76)}}  & \uline{4666.38$\pm$12.83}  &4703.48   & 495.11 & -5.39\%  \\
				M-Central100\_2            & 4730.11  & 4673.30  & 4784.78$\pm$93.23  &5008.39& 1743.75       & \cellcolor{gray}{\textbf{4479.99(-250.12)}}  & \uline{4556.14$\pm$28.18}  &4572.90   & 411.37 & -4.78\%  \\
				M-Central100\_3            & 4757.50  & 4527.97  & 4736.68$\pm$75.06  &4832.29& 1618.37       & \cellcolor{gray}{\textbf{4447.56(-309.94)}}  & \uline{4498.85$\pm$42.07}  &4589.85   & 738.21 & -5.02\%  \\
				M-Central100\_4            & 4697.35  & 4452.41  & 4604.47$\pm$115.32 &4792.95& 1036.62       & \cellcolor{gray}{\textbf{4257.75(-439.60)}}  & \uline{4351.95$\pm$25.09}  &4363.22   & 470.07 & -5.48\%  \\
				M-Central100\_5            & 4732.49  & 4681.76  & 4766.02$\pm$70.87  &4962.86& 1254.83       & \cellcolor{gray}{\textbf{4465.08(-267.41)}}  & \uline{4552.13$\pm$40.70}  &4580.99   & 445.71 & -4.49\%  \\
				M-Central100\_6            & 4524.44  & 4468.79  & 4528.98$\pm$52.75  &4695.14& 983.92        & \cellcolor{gray}{\textbf{4257.08(-267.36)}}  & \uline{4370.95$\pm$27.73}  &4381.80   & 527.36 & -3.49\%  \\
				M-Central100\_7            & 4786.57  & 4558.80  & 4786.81$\pm$93.81  &5021.92& 1166.55       & \cellcolor{gray}{\textbf{4462.69(-323.88)}}  & \uline{4559.14$\pm$29.47}  &4575.32   & 428.45 & -4.76\%  \\
				M-Central100\_8            & 4784.18  & 4507.96  & 4750.78$\pm$111.02 &5001.91& 923.39        & \cellcolor{gray}{\textbf{4436.81(-347.37)}}  & \uline{4450.38$\pm$6.40}   &4459.45   & 751.02 & -6.32\%  \\
				M-Central100\_9            & 4795.70  & 4705.49  & 4814.19$\pm$68.67  &5018.92& 1154.06       & \cellcolor{gray}{\textbf{4507.85(-287.85)}}  & \uline{4603.30$\pm$18.39}  &4611.93   & 403.65 & -4.38\%  \\
				M-Central100\_10           & 4686.57  & 4496.40  & 4605.44$\pm$102.27 &4763.41& 892.67        & \cellcolor{gray}{\textbf{4366.85(-319.72)}}  & \uline{4372.59$\pm$3.01}   &4377.55   & 477.49 & -5.06\%  \\
				\midrule
				No. Best / W-D-L & 0&  0       &                    &       &               & 30                                           & 30-0-0                     &          &        &          \\
				Mean & 2784.95  & 2718.07  & 2797.04$\pm$47.28  &2890.38& 644.51            & 2639.86                                      & 2672.70$\pm$12.32          &2686.09   & 218.30 & -3.98\%  \\
				
				\bottomrule
			\end{tabular}
			\label{tab:mid_scale}
		}
	\end{center}
\end{table*}

\subsection{Results on Small, Medium, and Large-Scale Instances}
\label{sec:small_mid_results}
Tables~\ref{tab:small_scale}-\ref{tab:large_scale} present the comparative results on the S-Central, M-Central, and Beijing benchmark sets, respectively.
A brief description of the contents in the tables is given below.
\begin{itemize}
    \item For both METS and GRASP, the column ``Best($\Delta$BKS)'' reports the best solution quality in terms of total distance (TD) across all runs, with the value in parentheses indicating the improvement over the previous best-known solution (BKS). The column ``Worst'' reports the worst solution quality in terms of TD.
    The column ``Avg$\pm$Std'' presents the average TD and its standard deviation.
    The ``Time'' column shows the average time-to-best.
    Since GRASP found most of the current BKSs for instances in the M-Central set, the best results reported in the original GRASP paper are included in Tables~\ref{tab:small_scale} and \ref{tab:mid_scale} for comparison, with columns being indicated by the corresponding citation~\cite{RN6219}.
    The column ``Imp. A'' reports the improvement ratio of METS over GRASP in terms of average solution quality, calculated as the difference between METS's result and GRASP's result, divided by GRASP's result.
    Thus, a negative ratio indicates that METS outperforms GRASP (i.e., achieves a lower solution cost), with larger absolute values representing greater improvements.
    Results are considered statistically significant when $p<0.05$, and these significantly better results are underlined in the tables.
    \item For CP-Proactive and Hexaly, the ``Best'' results under a 3600-second time limit are reported in Table~\ref{tab:small_scale}. 
    The results of CP-Proactive are obtained from the original paper~\cite{RN6217}, and the results of Hexaly are from our experiments.
    For Hexaly, we developed two formulations of GRVRP-PCAFS: a standard MILP model (as the Hexaly solver can directly process such models) and a Hexaly model using solver-specific features (e.g., lists, sets).
    The results of using the Hexaly model are reported in Table~\ref{tab:small_scale} since they are consistently better than the results of using the MILP model.
    The mathematical formulations and complete results of both models are provided in Appendix~D of the supplementary.
    Both CP-Proactive and Hexaly are excluded from Tables~\ref{tab:mid_scale} and~\ref{tab:large_scale} since they cannot find feasible solutions within the given time limit.
    \item In the last two rows of these tables, ``No. Best / W-D-L'' shows the number of instances in the benchmark set where an algorithm achieves the best solution quality. For the ``Avg$\pm$Std'' column, the ``No. Best / W-D-L'' row shows the win-draw-loss (W-D-L) counts of the statistical comparison between METS and GRASP, indicating the number of instances where METS performs significantly better, shows no significant difference, or performs significantly worse than GRASP.
    ``Mean'' presents the mean values across all instances for each metric.
    \item For each instance, the best solution quality among all ``Best'' columns is indicated in bold.
    \textbf{New} BKSs that improve upon the previous ones are highlighted in gray.
\end{itemize}

\begin{table*}[t]
\caption{Comparisons between GRASP and METS on large-scale instances. For each instance, the best performance is indicated in bold, new best-known solutions are highlighted in gray, and results significantly better than GRASP, based on 10 independent runs, are marked with underlines according to a non-parametric test at a significance level of p-values \( < \) 0.05.}
\begin{center}
\scalebox{0.88}{ 
\begin{tabular}{l 
                r l r r 
                l l r r 
                r }
\toprule
 \multirow{2}{*}{Instance}              
 & \multicolumn{4}{l}{ \multirow{2}{*}{GRASP}}                   
 & \multicolumn{4}{l}{ \multirow{2}{*}{METS}}                       
 & \multicolumn{1}{l}{Imp. A}   
\\ 
 & \multicolumn{4}{l}{ {}}                        
 & \multicolumn{4}{l}{ {}}                     
 & \multicolumn{1}{c}{(\%)}    
 \\ 
                            \cmidrule(lr){2-5}                        \cmidrule(lr){6-9}
                               &  Best  &  Avg$\pm$Std         & Worst    & Time (s)      & Best($\Delta$BKS)                               &  Avg$\pm$Std             &Worst    &Time (s)  &          \\ 
\midrule
Beijing200\_1                  & 4501.82  & 4625.52 $\pm$141.43 &4976.68  & 744.67        & \cellcolor{gray}{\textbf{4371.96(-129.86)}}  &  \uline{4471.99$\pm$63.05}  &4538.68  & 793.31   & -3.32\%  \\
Beijing200\_2                  & 6427.97  & 6518.65 $\pm$67.56  &6636.71  & 1715.03       & \cellcolor{gray}{\textbf{6334.57(-93.40)}}   &  \uline{6338.20$\pm$2.11}   &6341.53  & 468.90   & -2.77\%  \\
Beijing200\_3                  & 4534.42  & 4641.68 $\pm$121.55 &4969.47  & 859.02        & \cellcolor{gray}{\textbf{4408.75(-125.67)}}  &  \uline{4517.95$\pm$66.66}  &4568.91  & 1150.40  & -2.67\%  \\
Beijing200\_4                  & 6638.99  & 6661.33 $\pm$16.87  &6696.02  & 3059.21       & \cellcolor{gray}{\textbf{6407.07(-231.92)}}  &  \uline{6421.63$\pm$6.74}   &6429.51  & 957.12   & -3.60\%  \\
Beijing400\_1                  & 8555.42  & 8806.39 $\pm$161.84 &9058.96  & 2750.79       & \cellcolor{gray}{\textbf{8530.72(-24.70)}}   &  \uline{8540.35$\pm$3.71}   &8544.13  & 3599.28  & -3.02\%  \\
Beijing400\_2                  & 12561.23 & 12832.85$\pm$113.57 &12907.64 & 1338.38       & \cellcolor{gray}{\textbf{12277.58(-283.65)}} &  \uline{12363.36$\pm$58.91} &12428.45 & 5394.36  & -3.66\%  \\
Beijing400\_3                  & 8894.16  & 9078.44 $\pm$197.85 &9447.34  & 3554.22       & \cellcolor{gray}{\textbf{8759.47(-134.69)}}  &  \uline{8776.04$\pm$7.06}   &8784.84  & 2943.84  & -3.33\%  \\
Beijing400\_4                  & 12827.50 & 12859.35$\pm$17.44  &12877.33 & 5794.46       & \cellcolor{gray}{\textbf{12534.83(-292.67)}} &  \uline{12541.83$\pm$3.63}  &12546.97 & 3177.82  & -2.47\%  \\
Beijing600\_1                  & 12973.91 & 13107.71$\pm$108.85 &13253.20 & 3842.63       & \cellcolor{gray}{\textbf{12499.83(-474.47)}} &  \uline{12635.17$\pm$50.50} &12670.53 & 2642.80  & -3.61\%  \\
Beijing600\_2                  & 19045.21 & 19098.85$\pm$62.95  &19264.50 & 1713.12       & \cellcolor{gray}{\textbf{18216.44(-828.81)}} &  \uline{18230.19$\pm$5.46}  &18234.95 & 3030.78  & -4.55\%  \\
Beijing600\_3                  & 13480.33 & 13637.78$\pm$127.36 &13872.40 & 4030.21       & \cellcolor{gray}{\textbf{12971.00(-509.33)}} &  \uline{13004.98$\pm$43.14} &13121.15 & 3062.95  & -4.64\%  \\
Beijing600\_4                  & 19590.56 & 19624.95$\pm$48.64  &19659.35 & 3718.51       & \cellcolor{gray}{\textbf{18729.00(-861.56)}} &  \uline{18741.17$\pm$6.34}  &18749.74 & 3544.35  & -4.50\%  \\
Beijing800\_1                  & 17202.25 & 17278.42$\pm$94.50  &17459.38 & 4132.25       & \cellcolor{gray}{\textbf{16639.33(-562.92)}} &  \uline{16698.09$\pm$65.08} &16785.43 & 3249.93  & -3.36\%  \\
Beijing800\_2                  & 25256.34 & 25429.95$\pm$109.74 &25658.16 & 2619.32       & \cellcolor{gray}{\textbf{24339.56(-916.78)}} &  \uline{24350.28$\pm$7.03}  &24360.77 & 3654.35  & -4.25\%  \\
Beijing800\_3                  & 17693.31 & 17901.67$\pm$109.36 &18078.06 & 4555.37       & \cellcolor{gray}{\textbf{17160.56(-532.75)}} &  \uline{17251.10$\pm$74.50} &17330.79 & 2911.35  & -3.63\%  \\
Beijing800\_4                  & 25818.11 & 25905.35$\pm$123.38 &25992.59 & 2846.25       & \cellcolor{gray}{\textbf{24895.48(-922.62)}} &  \uline{25003.68$\pm$77.89} &25080.95 & 4323.42  & -3.48\%  \\
Beijing1000\_1                 & 21200.92 & 21434.34$\pm$131.48 &21629.02 & 4630.98       & \cellcolor{gray}{\textbf{20660.36(-540.56)}} &  \uline{20930.28$\pm$113.35}&21098.49 & 3564.84  & -2.35\%  \\
Beijing1000\_2                 & 31259.10 & 31413.30$\pm$78.26  &31571.31 & 5972.08       & \cellcolor{gray}{\textbf{30263.36(-955.74)}} &  \uline{30320.29$\pm$78.26} &30466.63 & 3017.30  & -3.48\%  \\
Beijing1000\_3                 & 21972.00 & 22204.79$\pm$119.05 &22380.07 & 2580.45       & \cellcolor{gray}{\textbf{21353.74(-618.26)}} &  \uline{21507.25$\pm$57.66} &21564.12 & 4614.13  & -3.14\%  \\
Beijing1000\_4                 & 32182.91 & 32201.06$\pm$12.13  &32221.09 & 6170.15       & \cellcolor{gray}{\textbf{31237.51(-945.40)}} &  \uline{31278.36$\pm$47.07} &31403.46 & 4872.71  & -2.87\%  \\
\midrule
No. Best / W-D-L &   0  &                     &         &               & 20                                           & 20-0-0                      &         &          &          \\
Mean     & 16130.82     & 16263.12$\pm$98.19  &16430.46 & 3331.36       & 15629.53                                     &  15696.11$\pm$41.91         &15752.50 &3048.70   & -3.43\%  \\
\bottomrule
\end{tabular}
\label{tab:large_scale}
}
\end{center}
\end{table*}

From Tables~\ref{tab:small_scale}-\ref{tab:large_scale}, the effectiveness of METS can be evaluated from two aspects: solution quality and robustness.
The former includes the best and average solution quality across instances, while the latter considers worst-case performance and solution stability over repeated runs.
Regarding best solution quality, METS obtains the best solutions on 57 out of 60 GrVRP-PCAFS instances, which is significantly better than the compared algorithms.
Specifically, for all 30 medium-scale instances (Table~\ref{tab:mid_scale}), METS discovers new BKSs across all of them over the previous ones found by GRASP.
The improvement ratios in these new BKSs over previous ones are considerable, with METS achieving an average improvement of 1.38\% compared to GRASP.
This observation extends to the 20 large-scale instances (Table~\ref{tab:large_scale}), where METS finds the best solutions across all instances with an even larger average improvement ratio of 3.90\% compared to GRASP.
Regarding robustness, METS yields better worst-case performance on most instances. It also shows lower standard deviations than GRASP, indicating more consistent and stable performance across runs.

For small-scale instances, METS achieves slightly fewer best solutions compared to CP-Proactive, which is reasonable given CP-Proactive's nature as an exact method.
However, in the original paper~\cite{RN6217}, CP-Proactive executes under a maximum runtime of 3600s, and solutions returned at this time limit are not guaranteed to be optimal due to early termination.
This situation occurs on instances S-Central\_3 and S-Central\_4.
Notably, for S-Central\_4, METS discovers a new BKS that improves upon the solution found by CP-Proactive.
In addition, the general-purpose global optimization solver Hexaly consistently found feasible solutions on small-scale instances within the  3600-second time limit.
Actually, it demonstrated strong performance, identifying six optimal solutions and achieving results close to BKSs for the remainder.
Nevertheless, Hexaly was still outperformed by METS, indicating the value of developing a specialized algorithm for GrVRP-PCAFS.
Compared to GRASP, METS not only finds more best solutions and achieves an average improvement of 1.38\% in best solution quality, but also demonstrates better robustness, as evidenced by its lower worst-case costs and smaller standard deviations across most instances.

In terms of average solution quality, METS demonstrates significantly better performance than GRASP on 59 out of 60 instances, with only one exception in the S-Central set.
The improvements are substantial across all problem sizes, with average improvement ratios of 1.59\%, 3.98\%, and 3.43\% for small, medium, and large-scale instances, respectively.
In terms of robustness, METS maintains tighter standard deviations than GRASP across nearly all instances, indicating greater consistency in solution quality.
Furthermore, METS consistently achieves better worst-case results than GRASP, suggesting its ability to avoid poor solutions even in challenging runs.
In summary, these results validate the effectiveness of METS, as it achieves improvements over existing methods in both best and average solution quality, by a large margin.
Finally, regarding time-to-best, METS and GRASP show comparable average results for small and large-scale instances, while METS demonstrates a clear advantage for medium-scale instances.
Considering that METS consistently finds solutions of higher quality than GRASP, it can be concluded that METS has higher search efficiency than GRASP.
This advantage may stem from the contribution of multiple algorithmic components, which will be further examined through ablation studies in Section~\ref{subsec:Ablation study}.

\subsection{Ablation study}
\label{subsec:Ablation study}

\begin{table*}
\caption{Results of variants and METS on small-scale and medium-scale instances. Each instance was solved 30 times and the algorithm terminates when the maximum number of iterations reaches 2000 or when there have been 300 consecutive iterations without improvement. The best performance is indicated in bold.}
\begin{center}
\scalebox{0.88}{ 
\begin{tabular}{l
                r r 
                r r r
                r r r
                r r r
                }
\toprule
 \multirow{1}{*}{Size}  
  
 & \multicolumn{3}{l}{ \multirow{1}{*}{METS-WCAI}}        
 & \multicolumn{3}{l}{ \multirow{1}{*}{METS-WFit}}
 & \multicolumn{3}{l}{ \multirow{1}{*}{METS-WAP}}
  & \multicolumn{2}{l}{ \multirow{1}{*}{METS}}   
\\ 
                                    \cmidrule(lr){2-4}                     \cmidrule(lr){5-7}                   \cmidrule(lr){8-10}                  \cmidrule(lr){11-12}
                                    & Avg(Deg\%)    & Time (s)& W-D-L   & Avg(Deg\%)    &Time (s)&W-D-L   & Avg(Deg\%)    &Time (s)& W-D-L  & Avg  & Time (s)          \\ 
\midrule
15                                  & 916.83(+3.41\%)  & 26.80   & 0-0-10  & 895.94(+1.15\%)  & 22.41  & 0-2-8  & 897.70 (+1.35\%) & 16.46  & 0-1-9  & \textbf{885.61}  & 15.69 \\ 
25                                  & 1177.53(+0.49\%) & 81.92   & 0-0-10  & 1178.20(+0.54\%) & 26.22  & 0-5-5  & 1180.37(+0.73\%) & 20.18  & 0-6-4  & \textbf{1171.78} & 20.25 \\ 
50                                  & 2407.07(+2.45\%) & 310.18  & 0-0-10  & 2352.15(+0.17\%) & 113.40 & 3-2-5  & 2362.55(+0.61\%) & 84.53  & 1-3-6  & \textbf{2348.14} & 119.82\\
100                                 & 4546.55(+1.06\%) & 1155.54 & 0-0-10  & 4505.92(+0.17\%) & 455.96 & 1-1-8  & 4525.87(+0.62\%) & 226.94 & 0-0-10 & \textbf{4498.18} & 514.84\\ 
\midrule
Mean              & 2262.00(+1.59\%) & 393.61  &         & 2233.05(+0.32\%) & 154.50 &        & 2241.62(+0.70\%) & 87.03  &        & \textbf{2225.93} & 167.65\\
\bottomrule
\end{tabular}
\label{tab:Ablation study}
}
\end{center}
\end{table*}

To investigate the individual contribution of each key component in METS, we construct five variants of METS, where each variant disables one component while retaining all others.
These variants are as follows:

\begin{itemize}
 \item\textbf{METS-WCAI:} it removes the CAI rule and adds a standard AFS insertion operator to preserve the ability to insert AFSs when necessary. This operator simply selects the position that yields the minimum additional distance within a route, without considering feasibility status.
 \item\textbf{METS-Wmove:} it removes all CAI-based move operators ($N_{1}$-$N_{4}$) from the local search procedure, and no standard AFS insertion operator is introduced in this variant.
 \item\textbf{METS-WFit:} it removes the use of the comprehensive fitness function and relies solely on total quality defined in Eq.~(\ref{eq:fit}).
 \item\textbf{METS-WAP:} it disables the adaptive penalty adjustment mechanism. Instead of updating penalty weights during the search process, this variant uses fixed penalty values from Table~\ref{tab1} throughout the entire search.
 \item\textbf{METS-Wfast:} it removes the constant-time move evaluation mechanism.
\end{itemize}

Due to the prohibitive computational time on large-scale instances, these experiments focused on instances with up to 100 customers, as the results already demonstrated clear benefits of these components.
For fair comparison, the first four variants (METS-WCAI, METS-Wmove, METS-WFit, METS-WAP) and METS shared the same termination criteria: 2000 iterations or 300 consecutive iterations without improvement. 
In contrast, since fast evaluation aims to accelerate the assessment of neighboring solutions, both algorithms were given a time limit of 10 times the number of customers as the termination condition. 
Hence, the ablation results are presented in two groups based on their respective stopping criteria.

Table~\ref{tab:Ablation study} presents the comparative results of METS and its variants in terms of average solution quality and time-to-best across 30 repeated runs on small-scale and medium-scale instances.
The ``Deg'' reports the degradation ratio of each variant relative to METS in terms of average solution quality. It is calculated as the difference between the variant's result and METS's result, divided by the variant's result. A positive ratio indicates that the variant performs worse than METS (i.e., yields a higher solution cost), with larger values representing greater degradation.
The ``W-D-L'' columns indicate the number of instances where the variant performed significantly better, showed no significant difference, or performed significantly worse than METS, respectively.
Notably, METS-Wmove is not included in the table, as it failed to find any feasible solution on all test instances.
In all cases, removing a component leads to a degradation in solution quality. 
Specifically, the average solution cost increased by 1.59\%, 0.32\%, and 0.70\% for METS-WCAI, METS-WFit, and METS-WAP, respectively, compared to METS. 
The W-D-L results further highlight the negative impact of component removal, especially for METS-WCAI, which performed significantly worse than METS on all instances. 
These findings indicate that among the key components of METS, the CAI rule contributes the most to overall performance. 
The fact that METS-Wmove failed to produce any feasible solution further reinforces the importance of this component.

\begin{figure}[tbp]
\centering
\includegraphics[width=\linewidth]{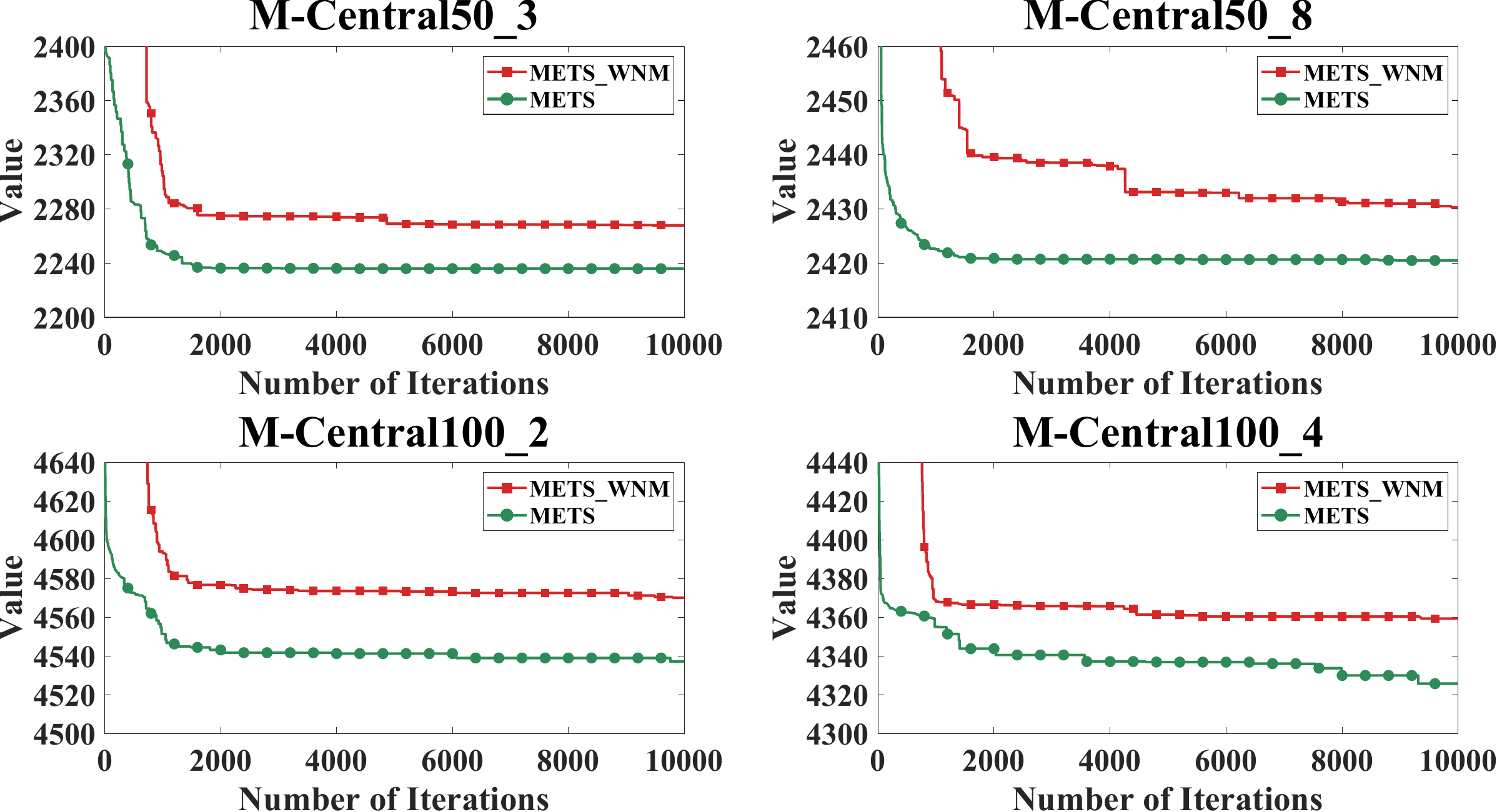}
\caption{The convergence profiles of METS-WNM and METS on the M-Central50\_3, M-Central50\_8, M-Central100\_2, and M-Central100\_4 instances across 30 independent runs. The lines represent the average best values with respect to the number of iterations.}
\label{fig:ls}
\end{figure}

For a deeper analysis, we ran both METS and METS-WCAI 30 times on four representative instances.
The convergence charts in Fig.~\ref{fig:ls} show that METS consistently outperforms METS-WNM across all iterations.
Both algorithms reduce the objective value quickly within the first 2000 iterations.
However, METS maintains a clear advantage throughout the process, delivering better solutions at every stage.

We next analyze the effect of the constant-time move evaluation by comparing METS and METS-Wfast under equal runtime budgets.
Table~\ref{tab:ls_fast} presents the comparative results following the same format as Table~\ref{tab:Ablation study}.
METS consistently outperforms METS-Wfast across all problem sizes, owing to its ability to evaluate more solutions within the same time. And as problem size increases, the performance gap widens.
Fig.~\ref{fig:fast} illustrates the average runtime per generation and the number of evaluated neighbors.
METS achieves lower per-generation runtime and explores more neighbors than METS-Wfast, with the advantage growing on larger instances. These results confirm the efficiency of the fast evaluation mechanism.
under increasing computational demands.

\begin{table}[tbp]
\caption{Results of METS-Wfast and METS on small-scale and medium-scale instances. Each instance was solved 30 times, the cut-off time was set to be ten times the number of customers. The better performance is indicated in bold.}
\begin{center}
\scalebox{0.88}{ 
\begin{tabular}{l 
                r r r
                r r}
\toprule
 \multirow{2}{*}{Size}  
 & \multicolumn{3}{l}{ \multirow{1}{*}{METS-Wfast}}                   
 & \multicolumn{2}{l}{ \multirow{1}{*}{METS}}                 
\\ 
                                \cmidrule(lr){2-4}                                \cmidrule(lr){5-6} 
                                & Avg(Deg\%) &  Time (s)   & W-D-L             &  Avg               & Time (s)  \\ 
\midrule
15                              & 893.41 (+0.16\%) & 33.17    & 0-10-0            & \textbf{\underline{891.94}}  & 13.59     \\
25                              & 1178.14(+0.09\%) & 46.88    & 0-10-0            & \textbf{1177.05}             & 22.37     \\ 
50                              & 2369.85(+0.46\%) & 221.62   & 0-6-4             & \textbf{\underline{2358.90}} & 144.38    \\ 
100                             & 4540.04(+0.58\%) & 481.43   & 0-4-6             & \textbf{\underline{4513.70}} & 435.68    \\ 
\midrule
Mean          & 2245.36(+0.44\%) & 195.78   &                   & 2235.40                      & 154.01    \\
\bottomrule
\end{tabular}
\label{tab:ls_fast}
}
\end{center}
\end{table}

\begin{figure}[t]
\centering
\includegraphics[width=1.0\linewidth]{fig/Fig_4.pdf}
\caption{The average ratio of visited solutions and the average runtime per generation of METS and METS-Wfast on nine instances of different sizes. Solid blue lines represent the runtime of METS, while dashed lines represent the runtime of METS-Wfast.}
\label{fig:fast}
\end{figure}

\subsection{A Deeper Look into the Dynamics of Diversity-Feasibility}
\begin{figure}[tbp]
\centering
\includegraphics[width=\linewidth]{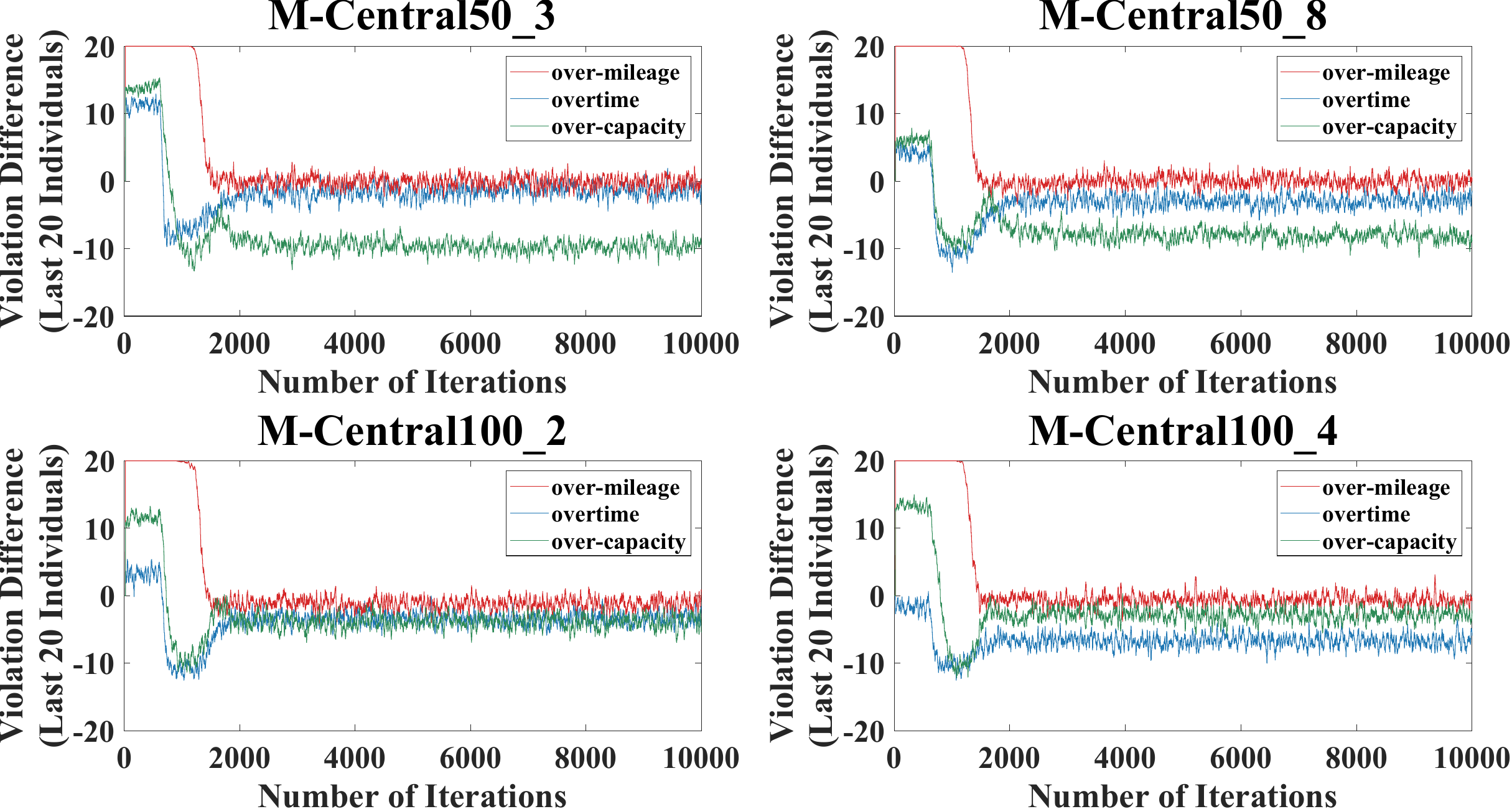}
\caption{The convergence profiles of METS on four representative instances in terms of feasibility violation difference. 
The lines represent the average of difference between the number of feasible and infeasible individuals among the last 20 individuals for each constraint type: overtime, over-mileage, and over-capacity. 
Positive values indicate more feasible individuals.}
\label{fig:fea}
\end{figure}

\begin{figure}[tbp]
\centering
\includegraphics[width=\linewidth]{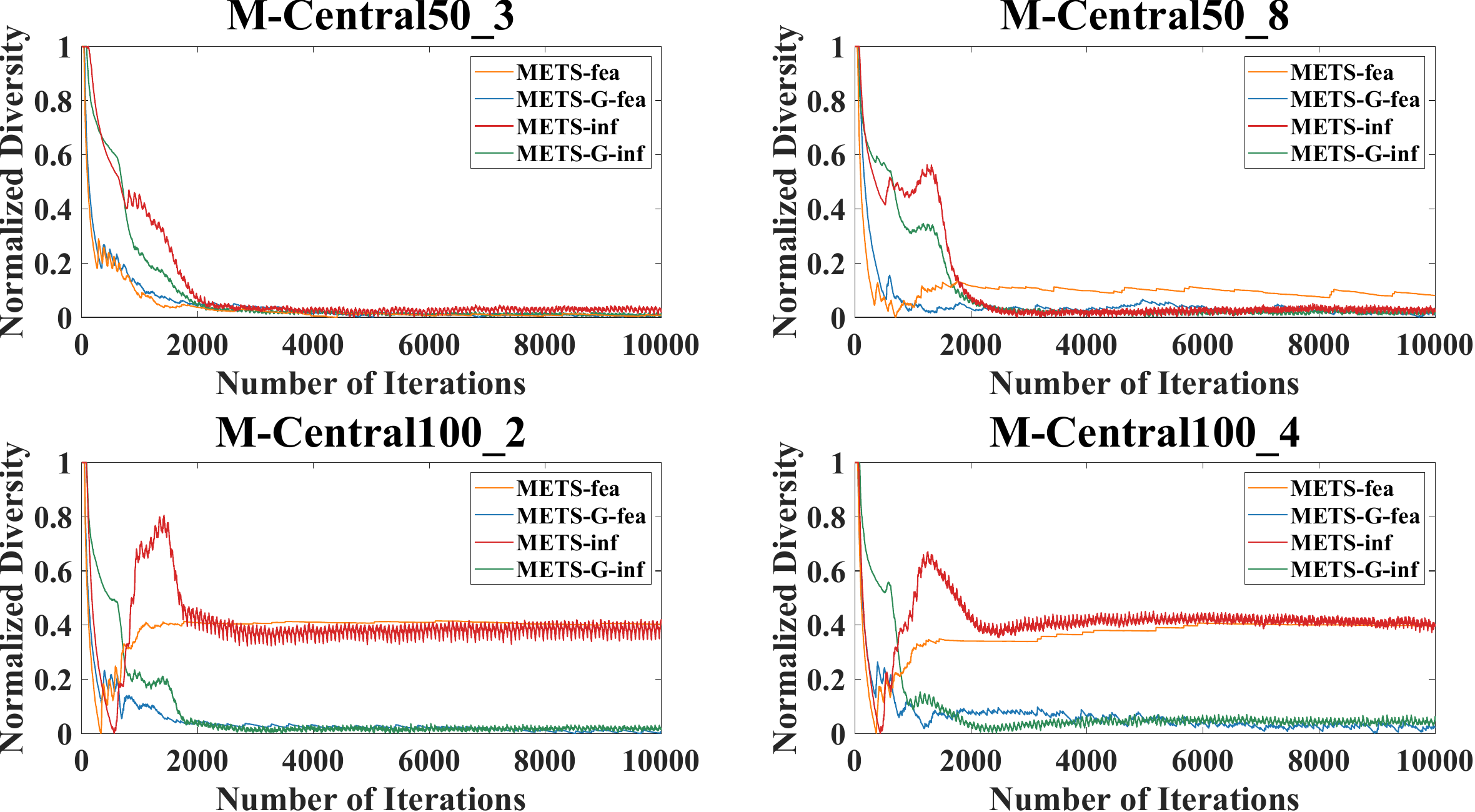}
\caption{The convergence profiles of METS and METS-G on four representative instances in terms of normalized average subpopulation diversity. The lines represent the average diversity of individuals in each subpopulation, normalized using min-max scaling within each instance. 
Each algorithm maintains two types of subpopulations: feasible (fea) and infeasible (inf). 
Higher values reflect greater population diversity during the search process.}
\label{fig:div}
\end{figure}

To analyze how METS balances diversity and feasibility, we examined its convergence profiles on four representative instances, focusing on the dynamics of both aspects.
Each instance was independently run 30 times, and the results are shown in Fig.~\ref{fig:fea} and Fig.~\ref{fig:div}, respectively.

To evaluate the effectiveness of the SCTS strategy in contribution of diversity, a variant algorithm, METS-G, was created by replacing SCTS with the construction phase of GRASP~\cite{RN6219}. 
Fig.~\ref{fig:div} shows the normalized diversity dynamics of feasible and infeasible subpopulations for METS and METS-G.
At each iteration, the diversity of a subpopulation is calculated as the average of individual diversity values, computed using Eq.~(\ref{eq:div}) and normalized via min-max scaling within each instance. 
METS consistently maintains higher diversity levels, especially in the infeasible subpopulation during the early and middle stages, demonstrating the advantage of SCTS in generating diverse offspring. 

To further support this design choice, additional experiments were performed, including probabilistic mixing of the two split criteria, a direct bi-constraint split enforcing both $D_{max}$ and $T_{max}$ simultaneously, and a comparison with METS-G where SCTS is replaced by the GRASP construction phase.
The detailed procedures, pseudocode, and results are provided in Appendix~C of the supplementary.

Fig.~\ref{fig:fea} presents the feasibility violation differences in the last 20 individuals across four representative instances. 
These individuals are selected to assess whether the adaptive adjustment of penalty weights and the comprehensive fitness function can effectively control constraint violations during exploration.
For each constraint type (i.e., over-mileage, overtime, and over-capacity), the violation difference represents the average of difference between the number of feasible and infeasible individuals. 
As shown in the charts, all three constraints are well controlled, with violation differences quickly converging and fluctuating around zero. 
This confirms that METS can guide the population efficiently toward feasible regions while maintaining overall constraint satisfaction.

In summary, the experimental results demonstrate that METS effectively maintains a balance between population diversity and solution feasibility throughout the search process.

\section{Conclusion}
\label{sec:conclusion}
In this work, a novel memetic algorithm, dubbed METS, is proposed to solve the Green Vehicle Routing Problem with Private Capacitated Alternative Fuel Stations (GrVRP-PCAFS).
METS incorporates three key components that strengthen the coordination between exploration and exploitation.
The effectiveness of METS has been validated through extensive experiments on both existing benchmark sets and a newly introduced large-scale benchmark set based on real-world logistics data.
Compared to existing approaches, METS discovered new best-known solutions on 31 out of 40 benchmark instances, achieving substantial improvements in solution quality.

Several promising directions for future research can be identified.
First, the proposed SCTS strategy, which generates diverse solutions by considering different constraints separately, combined with the fitness evaluation function for controlling diversity and feasibility, presents a new general idea of applying EAs to solve various VRP variants with complex constraints. Hence, we plan to explore this idea on other VRP variants beyond GrVRP-PCAFS.
Second, extending GrVRP-PCAFS into a multi-objective optimization framework could address practical needs by balancing travel time with service quality and fleet utilization while ensuring customer satisfaction~\cite{hanhuangMultiobj}.
Third, applying METS to more dynamic and complex logistics scenarios, particularly extra-large benchmark instances~\cite{lixiaodongLarge}, would enhance its effectiveness and scalability in real-world applications. 
Finally, given recent advances in using large language models (LLMs) for solving routing problems~\cite{liu1,KCTANllm}, investigating the integration of METS with LLMs for parameter recommendation and operator design presents an interesting research direction.

\section*{Acknowledgments}
This work was supported in part by the National Natural Science Foundation of China under Grant 62502192, in part by Hubei Provincial Natural Science Foundation of China under Grant 2024AFB338, and in part by the Fundamental Research Funds for the Central Universities under Grant B240207057.

\ifCLASSOPTIONcaptionsoff
  \newpage
\fi



\bibliographystyle{IEEEtran}
\bibliography{IEEEabrv, bibtex/bib/memeRE}

\vfill
\newpage
\includepdf[pages=-,pagecommand={}]{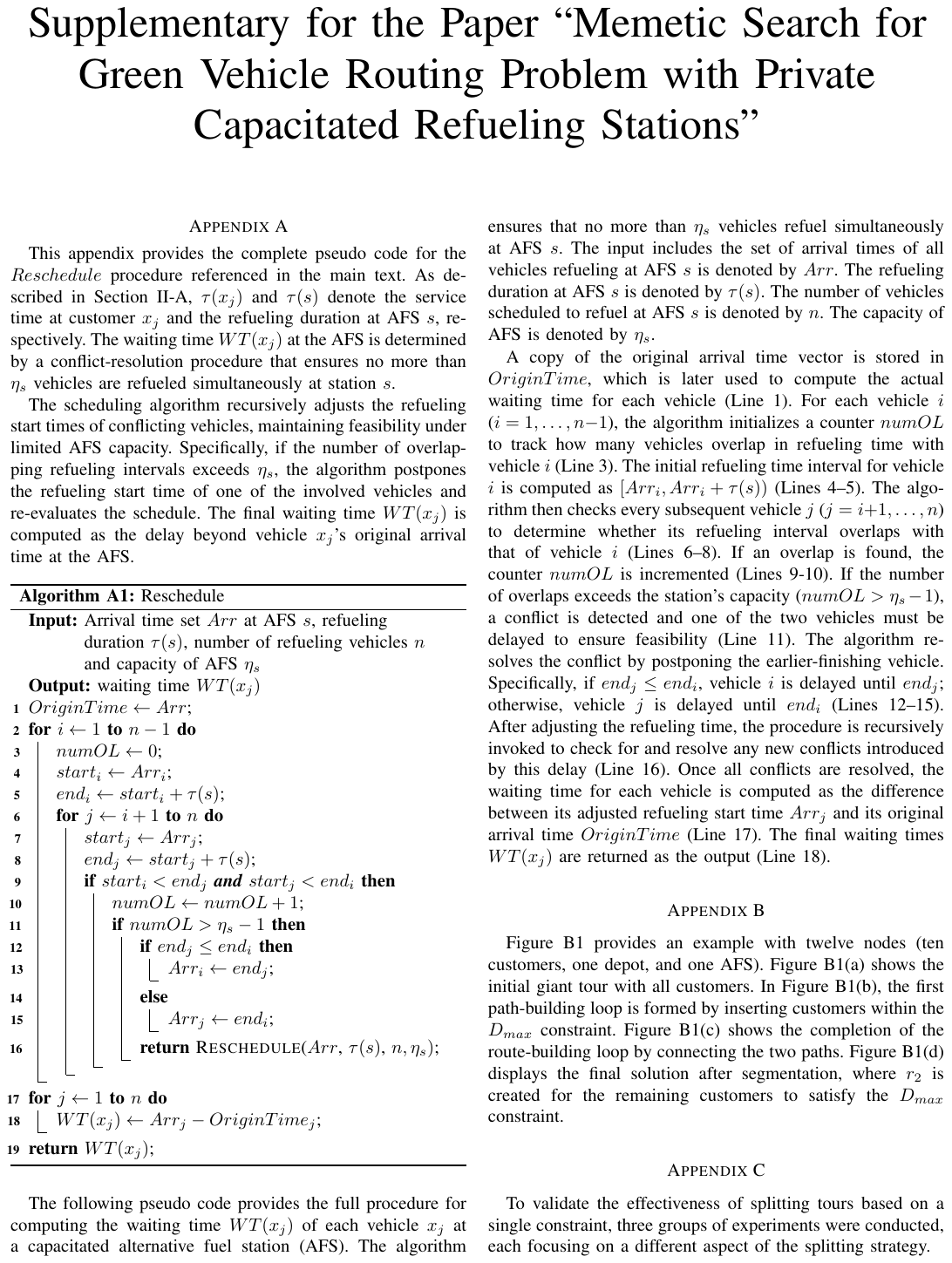}
\end{document}